\documentclass[journal,twoside]{IEEEtran}

\ifCLASSOPTIONcompsoc
  \usepackage[nocompress]{cite}
\else
  \usepackage{cite}
\fi

\ifCLASSINFOpdf
\else
\fi

\usepackage{algpseudocode}
\usepackage{algorithm}
\usepackage[fleqn]{amsmath}
\usepackage{amssymb,amsthm,amsfonts}
\usepackage{breqn}
\usepackage{mathtools}
\usepackage{cases}
\usepackage{mathrsfs}
\usepackage{breqn}
\usepackage{accents}
\usepackage{color}
\usepackage{tabu}
\usepackage{stfloats}
\usepackage{eucal}
\usepackage{mathptmx}
\usepackage{caption}
\usepackage{graphicx}
\usepackage{times}
\usepackage{array}
\usepackage{subfig}
\usepackage[T1]{fontenc}
\usepackage[latin1]{inputenc}
\usepackage{url}
\usepackage[normalem]{ulem}
\useunder{\uline}{\ul}{}
\usepackage{balance}
\usepackage{rotating}
\usepackage{epstopdf}
\usepackage{scrextend}
\usepackage{multirow}
\usepackage{siunitx}
\newlength{\dhatheight}

\newcommand{\etal}{\textit{et al.}}
\newcommand{\Fig}{Fig.}
\newcommand{\Figs}{Figs.}

\newtheorem{definition}{Definition}

\newcolumntype{P}[1]{>{\centering\arraybackslash}p{#1}}
\newcolumntype{M}[1]{>{\centering\arraybackslash}m{#1}}
\makeatletter
\newcommand*\bigcdot{\mathpalette\bigcdot@{1}}
\newcommand*\bigcdot@[2]{\mathbin{\vcenter{\hbox{\scalebox{#2}{$\m@th#1\bullet$}}}}}
\makeatother

\makeatletter
\renewcommand{\ALG@beginalgorithmic}{\small}
\makeatother
\newcommand{\subparagraph}{}
\usepackage{titlesec}

\titlespacing\section{-1pt}{12pt plus 4pt minus 2pt}{0pt plus 2pt minus 2pt}
\titlespacing\subsection{0pt}{12pt plus 4pt minus 2pt}{0pt plus 2pt minus 2pt}

\begin{document}


\title{A Scalable Framework for Trajectory Prediction}

\author{Punit~Rathore, Dheeraj~Kumar,~Sutharshan~Rajasegarar, ~Marimuthu~Palaniswami,~\IEEEmembership{Fellow,~IEEE},~James. C. Bezdek,~\IEEEmembership{Life Fellow,~IEEE}
\IEEEcompsocitemizethanks{\IEEEcompsocthanksitem Punit Rathore, and Marimuthu Palaniswami are with the Department of Electrical and Electronic Engineering, The University of Melbourne, Parkville, Victoria, Australia.
\protect\\
E-mail: \{prathore@student., palani@\}unimelb.edu.au.
\IEEEcompsocthanksitem Dheeraj Kumar is with the Lyles School of Civil Engineering, Purdue University, USA.
\protect\\
E-mail: kumar299@purdue.edu
\IEEEcompsocthanksitem Sutharshan Rajasegarar is with the School of Information Technology, Deakin University, Geelong, Victoria, Australia.
\protect\\
E-mail: srajas@deakin.edu.au.
\IEEEcompsocthanksitem James C. Bezdek is with the School of Computing and Information
Systems, The University of Melbourne, Victoria, Australia.
\protect\\
E-mail: jcbezdek@gmail.com.
}
}

%
%

\markboth{IEEE TRANSACTIONS ON Intelligent Transport Systems}
{Rathore \MakeLowercase{\textit{et al.}}: A Scalable Framework for Trajectory Prediction}
%


\IEEEtitleabstractindextext{%
\begin{abstract}
\textit{Trajectory prediction} (TP) is of great importance for a wide range of location-based applications in intelligent transport systems such as location-based advertising, route planning, traffic management, and early warning systems. In the last few years, the widespread use of GPS navigation systems and wireless communication technology enabled vehicles has resulted in huge volumes of trajectory data. The task of utilizing this data employing spatio-temporal techniques for trajectory prediction in an efficient and accurate manner is an ongoing research problem. Existing TP approaches are limited to short-term predictions. Moreover, they cannot handle a large volume of trajectory data for long-term prediction. To address these limitations, we propose a scalable clustering and Markov chain based hybrid framework, called Traj-clusiVAT-based TP, for both short-term and long-term trajectory prediction, which can handle a large number of overlapping trajectories in a dense road network. Traj-clusiVAT can also determine the number of clusters, which represent different movement behaviours in input trajectory data. In our experiments, we compare our proposed approach with a mixed Markov model (MMM)-based scheme, and a trajectory clustering, NETSCAN-based TP method for both short- and long-term trajectory predictions. We performed our experiments on two real, vehicle trajectory datasets,  including a large-scale trajectory dataset consisting of $3.28$ million trajectories obtained from $15,061$ taxis in Singapore over a period of one month. Experimental results on two real trajectory datasets show that our proposed approach outperforms the existing approaches in terms of both short- and long-term prediction performances, based on prediction accuracy and distance error (in km).
\end{abstract}

\begin{IEEEkeywords}
Large-scale trajectory data, next location prediction, Long-term trajectory prediction, scalable clustering.
\end{IEEEkeywords}}

\maketitle

\IEEEdisplaynontitleabstractindextext

%
\IEEEpeerreviewmaketitle


\section{Introduction}\label{sec:introduction}

With the widespread use of \textit{Global Positioning System} (GPS) devices, smart-phones,  Internet of Things~\cite{rathore2018real}, and wireless communication technology, it is possible to track all kinds of moving objects all over the world. The increasing prevalence of location-acquisition technologies has resulted in large volumes of spatio-temporal data, especially in the form of trajectories. These data often contain a great deal of information~\cite{zheng2008understanding}, which give rise to many location-based services (LBSs) and applications such as vehicle navigation, traffic management, and location-based recommendations. One key operation in such applications is the route prediction of moving objects.

Vehicle route prediction allows certain services to improve their quality e.g. if the route of vehicles is known in advance, \textit{intelligent transportation systems} (ITSs) can provide route-specific traffic information to drivers such as forecasting traffic conditions and routing the driver so as to avoid traffic jams. Route prediction also enables location-based advertising, which can advertise certain products/services and special offers to the target commuters most likely to pass through business outlets and stores based on their travel trajectory.

With the advancement in artificial intelligence, various machine learning techniques have now been used for trajectory analysis. Recently, several studies have been carried out on trajectory prediction, particularly after Song~\etal~\cite{song2010limits} demonstrated a 93\% potential for predictability in user mobility, which supplied the theoretical basis for location prediction methods. These methods mainly focus on two kinds of prediction models. The first type is the short-term trajectory prediction model, which aims to predict the next-location or a few locations in the near future. These models usually rely on current location and one or two previous locations of an object to predict its next location. The second type is the long-term trajectory prediction model which focuses on location prediction at a more distant future time or on complete route prediction. These models generally rely on an available partial trajectory of a moving object to predict the complete trajectory.

In urban areas, vehicle trajectories are usually constrained to a complex road network with many parallel and perpendicular road segments and intersections, which makes their time progression very irregular. Due to the uncertainty of moving objects, most of the existing trajectory prediction methods only focus on predicting short-term partial trajectories. They have poor prediction accuracy for long-term trajectory predictions and they do not work well for estimating continuous and complete trajectories.

The sheer amount of vehicle trajectory data, if analyzed effectively, can significantly improve route prediction performance. However, it is challenging to carry out trajectory prediction from a large amount of trajectory data. The huge volumes of data to be processed precludes using machine learning based \textit{traditional prediction} (TP) methods. Most of the existing trajectory prediction schemes cannot handle a large number of trajectories, especially when they span a large area of a road network.  Existing TP methods are hybrid in nature and usually employ classical frequent sequential pattern based algorithms, Markov model-based algorithms, or clustering based algorithms to a small amount of training data. Apriori-like sequential pattern methods bear significant computational overhead in mining frequent patterns because they produce a huge number of candidate sequences and require multiple scans of the database. Markov model approaches such as higher-order Markov models incur significant overhead in terms of computation time and space for storing and learning the mobility model. Similarly, parameter learning algorithms in Hidden Markov models impose significant computational burdens for large-scale trajectory datasets. Existing trajectory clustering based location prediction approaches are not scalable due to distance matrix computations for big data and are not able to handle a large number of overlapping trajectories efficiently. Therefore, most TP methods demonstrated in the literature use synthetic or small to medium size real trajectory datasets. To the best of our knowledge, the largest real dataset was used  in~\cite{lv2017big}, consisting of $370$ million GPS points. These authors utilized a parallel processing model, MapReduce, in their implementation to handle large datasets.

To address these challenges and overcome the drawbacks of existing TP methods, this paper presents a novel, scalable framework for both short-term and long-term TP, which is suitable for large numbers of overlapping trajectories in a dense road network, typical for major cities around the world. First, we cluster the large trajectory data using a modified version of our two-stage clusiVAT (\underline{clus}tering using \underline{i}mproved  \underline{V}isual \underline{A}ssessment of \underline{T}endency) algorithm~\cite{kumar2016hybrid,kumar2018fast}, which we call \textit{Traj-clusiVAT}, implemented for trajectory prediction task. The Traj-clusiVAT algorithm first extracts a smart sample using the \textit{Maximin-Random sampling} (MMRS) scheme~\cite{bezdek2017book}, which provides a good representation of the input cluster structure (present in the original data). Then, it uses the \textit{improved visual assessment of cluster tendency}  (iVAT) algorithm to visually determine the number of clusters ($k$) in input data, and subsequently, it partitions the trajectory sample into $k$ clusters which contain different frequent movement patterns in the trajectory data. Then, the remaining non-sampled trajectories are assigned to one of $k$ clusters using the \textit{nearest prototype rule} (NPR). Finally, Markov chain models are constructed from the trajectories in each cluster. These models quantify the movement patterns within clusters, and subsequently, can be used for TP.

Our main contributions in this paper are as follows:
\begin{itemize}
\item We propose a novel, hybrid framework for short-term and long-term trajectory prediction, based on a scalable clustering algorithm and Markov chain model, which can utilize a large number of trajectories in a dense road network.
\item We implement a modified version of the two-stage clusiVAT algorithm, called as Traj-clusiVAT, to cluster large numbers of trajectories in an accurate and efficient manner. In this regard,
\begin{itemize}
\item We develop a new algorithm to compute a \textit{representative trajectory} (RT) for each cluster, and subsequently, use it to assign non-sampled trajectories to one of the $k$ clusters in the nearest prototyping phase.
\item We also develop an improved algorithm for nearest prototyping, which assigns each non-sampled trajectory to one of the $k$ clusters, based on pattern matching and the distance of non-sampled trajectories to the RT of each cluster.
\end{itemize}
\item We compare our proposed algorithm with two TP algorithms: a \textit{mixed Markov model} (MMM)~\cite{asahara2011pedestrian} and a trajectory clustering model NETSCAN~\cite{kharrat2008clustering}  based algorithm. Our experiments used two real-life, taxi trajectory datasets including a large-scale taxi trajectory dataset consisting of $370$ million GPS traces and $3.28$ million passenger trips from $15,061$ taxis during the period of one month in Singapore. To the best of our knowledge, this
is the first time TP  has been performed on such a large number of real-life road network trajectories.
\end{itemize}

Here is an outline of the rest of this article. Section~\ref{relatedwork} discusses related work and limitations. Section~\ref{preliminaries} formally defines the problem and presents preliminaries on techniques used in our work.  Our novel, hybrid framework, based on Traj-clusiVAT and Markov models, is presented in Section~\ref{proposedframework}, and its time complexity is discussed in Section~\ref{sec:computationalcomplexity}. Section~\ref{sec:experiments} presents our numerical experiments, including computation protocols, datasets and results, followed by conclusions in Section~\ref{sec:conclusions}.
\section{Related Work}\label{relatedwork}
Several studies address the problem of trajectory predictions, which includes the problem of short-term prediction such as predicting the next location, and long-term prediction such as future locations or complete route prediction. These methods mainly focus on discovering frequent patterns using various data mining methods. Many of these methods are hybrid in nature and can be broadly classified into three categories: (i) Rule-based learning based approaches (ii) Markov model-based approaches (iii) Clustering-based approaches.
\subsection{Rule-based learning based approaches}\label{section2.1}
Several rule-based methods have been used for location prediction. Morzy~\cite{morzy2007mining} implemented a modified version of the PrefixSpan algorithm to extract association rules from a moving object database, and used frequent pattern tree with a matching function to select the best association rule from the database of movement rules. Jeung~\etal~\cite{jeung2008hybrid} proposed a hybrid prediction approach, which combines association rules in the form of trajectory patterns with the motion functions of an object's recent movements, to estimate future locations. Given an object's recent movement and predictive queries, the best association rule is chosen for prediction. The query processing approaches presented in~\cite{jeung2008hybrid} can only support near and distant-time predictive queries,  making them unsuitable for long-term trajectory prediction. Moreover, with the huge number of trajectories, the number of association rules is also huge, which makes association-rule based algorithms impractical for large-scale mobility data.

Monreale~\etal~\cite{monreale2009wherenext} built a decision tree that they called a T-pattern Tree, based on the frequent movement patterns extracted using a \textit{Trajectory Pattern} algorithm, and predicted the next location of a new trajectory based on the best matching functions. However, mining of frequent trajectory patterns is computationally expensive. The method in~\cite{monreale2009wherenext} is similar to the use of association rules as predictive rules in rule-based classifiers. Therefore, this method~\cite{monreale2009wherenext} may result in a large number of predictive rules for voluminous trajectories. Qiao~\etal~\cite{qiao2017predicting} proposed a TP algorithm, called PrefixTP, which examines only the prefix subsequences, and projects their corresponding postfix subsequences into projected sets. Then, for a partial trajectory, it recursively finds a postfix sequence based on the minimum support count requirement and then declares the most frequent sequential pattern as the most probable trajectory. Finding subsets of trajectory sequential patterns is a recursive mining process, which is also computationally extensive.

\subsection{Markov model-based approaches}\label{section2.2}
\textit{Markov models} (MMs) have been widely used to mine frequent patterns for route prediction problems. Ishikawa~\etal~\cite{ishikawa2004extracting} proposed a model to extract mobility statistics, called the Markov transition probability, which is based on a cell-based organization of target space and a Markov chain model, and employed R-tree spatial indices to compute Markov transition probabilities. Simmon~\etal~\cite{simmons2006learning} presented a \textit{Hidden Markov Model} (HMM) based probabilistic approach to predict a driver's intended route and destination through observations of the driver's habits. Asahara~\etal~\cite{asahara2011pedestrian} suggested that standard MM and HMM are not generic enough to encompass all types of movement behaviour. They proposed a variant of Markov model, called the \textit{mixed Markov-chain model} (MMM), as an intermediate model between individual and generic models, for pedestrian movement prediction.

Gambs~\etal~\cite{gambs2012next} extended a previously proposed mobility model, named $v$-\textit{Mobility Markov Chain} ($v$-MMC), to incorporate the $v$  previous visited locations. They showed that prediction accuracy increases with $v$, but increasing $v$ beyond two does not compensate for the significant overhead in terms of computation and space for learning and storing the mobility model. They only considered the sequence of the significant locations, instead of all locations, to build higher order MM.  \textcolor{black}{Zhang~\etal~\cite{zhang2016gmove} proposed a group-level mobility modeling method, Gmove, which alternates between two intertwined tasks, user grouping and mobility modeling, and generates an ensemble of HMMs to characterize group-level movement.} 

Most of the MMs do not consider the discontinuous chain of the hidden states, and therefore, the state retention problem can drastically degrade the accuracy of location prediction system~\cite{qiao2017predicting}. For the irregular trajectory data, the movement rules cannot be easily represented by Markov models, which may cause loss of continuous location information~\cite{qiao2017predicting}. Moreover, the HMM approaches use the Baum-Welch algorithm for parameter learning and the Viterbi algorithm to find the most likely sequences of hidden states. These algorithms impose a significant computation burden for large-scale trajectory datasets.

\textcolor{black}{Recently, deep learning techniques such as \textit{inverse reinforcement learning} (IRL), \textit{long short-term memory} (LSTM), and \textit{recurrent neural networks} (RNNs) have been used for modeling vehicle trajectories~\cite{altche2017lstm,wu2017modeling,bock2017self}. However, some of these studies are still based on first-order Markov assumption to model the routing decisions for heterogeneous destinations, or they are too shallow which makes the modeling pattern varieties suffer from too few parameters.}

\subsection{Clustering based approaches}\label{section2.3}

Some researchers have proposed trajectory clustering based route prediction methods, which partition the trajectories into several clusters representing different motion patterns based on the trajectory similarity. Various clustering approaches~\cite{yuan2017review} using different methods and distance measures between trajectories have been proposed in the literature. Road network constrained trajectory clustering approaches can be classified into two broad categories. The first type uses the traditional clustering approaches such $k$-means and DBSCAN with specially designed distance measures~\cite{won2009trajectory,lee2007trajectory,wang2012clustering,roh2010nncluster}  for trajectories.  The second category of algorithms~\cite{kharrat2008clustering,han2015road} cluster road segment vehicle frequencies based on density and flow.

Ashbrook~\etal~\cite{ashbrook2003using} presented a system that automatically detected the significant locations from GPS data using $k$-means clustering,  and then incorporated these locations into an MM to predict the next location. Mathew~\etal~\cite{mathew2012predicting} presented a hybrid method for human mobility prediction, which first clusters location histories according to their characteristics, and then trains an HMM for each cluster. A poor prediction accuracy of $13.85$\% was obtained on a real, large-scale trajectory dataset using this method. Chen~\etal~\cite{chen2015predicting} proposed a next-location prediction approach combining two clustering models, which cluster the objects based on the spatial locations and trajectories using a similarity metric, respectively, and trains a series of MMs with trajectories in each cluster.

Ying~\etal~\cite{ying2011semantic} proposed an approach for predicting the next location based on geographic and semantic features of user trajectories.  This method requires the calculation of a semantic score for each candidate path, which generally incurs additional
overhead when compared with other methods. A probabilistic TP model was proposed in~\cite{wiest2012probabilistic} based on two mixture models, a \textit{Gaussian Mixture Model} (GMM) and a \textit{Variational Gaussian Mixture Model} (VGMM), optimized using the \textit{Expectation Maximization} (EM) algorithm. Their method requires the prior selection of the number of Gaussian components and other distribution parameters. They evaluated their method on a small dataset, which consists of only  $69$ trajectories. Qiujian~\etal~\cite{lv2017big} proposed a spatio-temporal prediction and a next-place prediction model based on an entropy-based clustering approach and HMMs.

Traditional clustering~\cite{won2009trajectory,lee2007trajectory,wang2012clustering,roh2010nncluster} based prediction methods are not scalable for a large number of trajectories as distance matrix computation is time and space prohibitive. Most of them require the number of clusters to be known in advance, but in practice, it is often unknown, making it difficult for the user to choose the optimal number of clusters for location prediction. Furthermore, the clusters are determined by fixed rules. Some of the road network based clustering approaches~\cite{kharrat2008clustering,han2015road}, though scalable, produce clusters having high intra-cluster variance, which span a large area of a road network.

Most of the work done in the area of trajectory prediction either use synthetic datasets~\cite{morzy2007mining,jeung2008hybrid,asahara2011pedestrian,lei2011exploring} or real datasets with small to medium numbers of data points~\cite{chen2010system,monreale2009wherenext,chen2015predicting}. Most of them cannot handle big trajectory datasets.
There have been several attempts to demonstrate trajectory prediction on real data having a large number of samples. For example,~\cite{qiao2017predicting} uses a real dataset consisting of $4.9$ million trajectories ($790$ million GPS points) as a population, but only small subsets having a maximum $30,000$ trajectories are used in their experiments. To the best of our knowledge, the largest real dataset used was in~\cite{lv2017big}, consisting of  $37$ million GPS points. They utilized~\cite{lv2017big} the MapReduce model in their implementation to handle large datasets. In this paper, we use the GPS traces of $15,061$ taxis in Singapore over a period of one month. We extract $3.28$ million passenger trajectories consisting of $370$ million GPS logs from this data for trajectory prediction task. To the best of our knowledge, this is the first time trajectory prediction task has been performed on such a large number of real-world road network trajectories.

\section{Preliminaries}\label{preliminaries}
In this section, we introduce some basic terms and definitions, which are required in the sequel.

\subsection{Road Network and Trajectories}\label{section3.1}
We represent the road network as an undirected graph
\begin{eqnarray}
G_{RN}= (V,E),
\end{eqnarray}
comprising a set $V$ of \textit{intersections} or \textit{nodes} of the road network with a set $E$ of road \textit{segments} or \textit{edges}, $R_{i} \in E$ such that $R_{i}=(r_{i_{a}},r_{i_{b}})$, where $r_{i_{a}},r_{i_{b}} \in V$ and there exists a road between $r_{i_{a}}$ and $r_{i_{b}}$. The edge $R_{i}$ is given a weight equal to the length of $R_{i}$. For such a road network, we define the following:

\begin{definition}{(\textbf{Trajectory})}:
A \textbf{trajectory} $T$ of length $l$ is a time ordered sequence of \textit{road segments} (RS), $T =\langle R_{1},R_{2},...,R_{l}\rangle$, where $R_{j} \in E, 1\leq j\leq l$, and $R_{j}$ and $R_{j+1}$ are connected.
\end{definition}

\begin{definition}{(\textbf{Sub-Trajectory})}:
$T^{s} = \langle L_{1},L_{2},..,L_{p} \rangle$ is a \textbf{sub-trajectory} of sequence $T = \langle R_{1},R_{2},..,R_{l}\rangle$, $p \leq l$, if there are integers $\langle i_{1},i_{2},..i_{p} \rangle$ $(1 \leq i_{1}<i_{2}<...<i_{p})$, $\langle j_{1},j_{2},..j_{p} \rangle$ $(1 \leq j_{1}, j_{2}= (j_{1}+1),...,j_{p}=(j_{p-1}+1) \leq l)$, and $i_{1} \leq j_{1}$, $L_{i_{1}} =R_{j_{1}}$, $L_{i_{2}} =R_{j_{2}},.., L_{i_{p}} =R_{j_{p}}$. Then $T^{s}$ is called a sub-trajectory of $T$
, denoted by $T^{s} \sqsubseteq T$.
\end{definition}

\begin{definition}{(\textbf{Frequent Road Segment})}:
A \textbf{Frequent road segment} (FRS), $R_{FRS}$, in a trajectory set is a segment that contains at least $MinT$ percentage of trajectories of the set passing through the segment, otherwise, the segment is labeled as "non-FRS". The percentage $MinT$ is a tunable parameter and we call it the FRS threshold.
\end{definition}

\begin{definition}{(\textbf{Partial Trajectory})}:
A \textbf{partial trajectory} $T^{p}$ is a sub-trajectory of a given trajectory $T$ if and only if their sequences start from the same segment.
\end{definition}

\begin{definition}{(\textbf{Source Segment})}:  The segment from which a trajectory $T$ originates is called the \textbf{Source Segment} (SS), $R_{SS}$, and the start node of $T$ is called the \textbf{Source Node} (SN) of that trajectory. For a trajectory $T =\langle R_{1},R_{2},...,R_{l}\rangle$, the road segment $R_{1}$ is $R_{SS}$. Node $r_{1_{a}}$ is the SN, if $R_{2}$ has node $r_{1_{b}}$, else $r_{1_{b}}$ is SN, where $R_{1}=(r_{1_{a}},r_{1_{b}})$, and $r_{1_{a}},r_{1_{b}} \in V$.
\end{definition}

\begin{definition}{(\textbf{Frequent Source Segment})}:  The SS  which is FRS, is called \textit{Frequent source segment} (FSS), $R_{FSS}$.
\end{definition}

\begin{definition}{(\textbf{Problem Definition})}:  Assume that a historical trajectory database, containing $N$ trajectories, denoted by $\mathcal{T}= \{T_{1}, T_{2},...,T_{N}\}$ is given. Then, for a given partial trajectory $T^{p} = \langle L_{1},L_{2},...,L_{m}\rangle $, the goal is to predict the future road segments $L_{i}$, where $i,m \in \mathbb{Z}$ and $i \geq m+1$.
\end{definition}

\begin{table}[]
\centering
\caption{Notations}
\label{Table1}
\scalebox{0.9}{
\begin{tabular}{|c|l|}\hline
\textbf{Symbol} & \textbf{Definition} \\ \hline
$\mathcal{T}$ &  The set of trajectories \\ \hline
$T_{i}$ & The $i^{th}$ trajectory of set $\mathcal{T}$ \\ \hline
$l_{i}$ & The length (or number of segments)  of trajectory $T_{i}$ \\ \hline
$R_{i}$ & The $i^{th}$ segment of trajectory $T_{i}$ \\ \hline
$N$, $n$ & number of trajectories in $\mathcal{T}$ and MMRS sample $S$, respectively  \\  \hline
$k, K$ & number of non-directional and directional clusters in $S$ \\ \hline
$\mathcal{T}^{j}$ & Set of trajectories in cluster $j$ \\ \hline
$N_{j}$ & Number of trajectories in cluster $j$ \\ \hline
$\mathcal{R}^{j}$ & Set of points (segments) in cluster $j$ \\ \hline
$\mathcal{C}(\mathcal{T})$ & Set of cluster of trajectories \\ \hline
$R_{FRS}$, $\mathcal{R}_{FRS}$ & Frequent road segment (FRS) and the set of FRSs, respectively \\ \hline
$R_{SS}$, $\mathcal{R}_{SS}$ & Source segment and the set of SSs, respectively \\ \hline
$R_{FSS}$, $\mathcal{R}_{FSS}$ & Frequent source segment (FSS) and the set of FSSs, respectively \\ \hline
$M^{j}$ & Transition probability matrix for cluster $j$ \\ \hline
$W^{j}$ & Transition count matrix for cluster $j$ \\ \hline
\end{tabular}}
\end{table}
\vspace*{-1.8\baselineskip}
\subsection{Distance Measure (trajDTW)}\label{section3.2}
Most of the existing distance measures for trajectory similarity are not suitable for a large number of overlapping trajectories in a dense road network due to the use of either the number of overlapping road segments or maximum/minimum distance between trajectories in their computation. In our work, we use the Dijkstra based \textit{dynamic time warping} (DTW) distance measure, trajDTW~\cite{Kumar2015ASF} to compute trajectory similarities which is suitable for a large number of overlapping trajectories in a dense road network. The superiority of the trajDTW over the traditionally used \textit{dissimilarity with length} (DSL) and Hausdorff distance measures is demonstrated in~\cite{Kumar2015ASF}.  The trajDTW is a normal DTW algorithm with a Dijkstra distance matrix based cost function and a window parameter $w$, which is set to the half of the length of shorter of two trajectories, to avoid overestimation of the actual distance. As the road network is static, the distance matrix $D_{all}$ (of size $(|E| \times |E|)$ of all the edges $E$ in $G_{RN}$ can be pre-computed and stored.
\subsection{Non-directional trajDTW}\label{NonDirectTrajDTW}
The directionality of trajectories can result in misleading distances among them, which in turn may cause incorrect clustering results. For example, suppose there are two trajectories $T_{1}$ and $T_{2}$, which follow the same route but in opposite directions, then the distance between them considering their directions in computation will be higher than the distance computed without considering their directions. Therefore, if their movement direction is considered as part of the distance computation, $T_{1}$ and $T_{2}$ may not be grouped in the same cluster. The problem of incorrect distance measure due to the movement direction of trajectories is addressed by reversing one of them (reversing the sequence order so that the starting point becomes the ending point and vice versa), and taking the minimum distance between the first and second trajectory, and the first trajectory and second reverse trajectory. This distance is called  non-directional trajDTW~\cite{Kumar2015ASF}, and is given as:
\begin{dmath}\label{Eq:nondirectional_trajDTW}
\textrm{non-directional trajDTW}(T_{1},T_{2})=  \min (trajDTW(T_{1},T_{2}), trajDTW(T_{1},Reverse(T_{2}))
\end{dmath}
\subsection{The clusiVAT Algorithm}\label{clusiVATAlgo}
In our proposed framework, we modify the clusiVAT algorithm and use it for efficiently clustering large volumes of trajectory data.  The clusiVAT model finds its root in the \textit{visual assessment of cluster tendency} (VAT)~\cite{bezdek2002vat} and \textit{improved VAT} (iVAT)~\cite{havens2012efficient} algorithms. The VAT and iVAT algorithms reorder the input dissimilarity matrix $D_{N}$ (for $N$ datapoints) to $D_{N}^{*}$ (in VAT) and ${D'}_{N}^{*}$ (in iVAT), respectively, using a modified Prim's algorithm (for finding a \textit{minimum spanning tree} (MST)) and by applying a graph theoretic distance conversion, such that the dark blocks along the diagonal  of the reordered image $I(D_{N}^{*})$ (or  $I({D'}_{N}^{*})$) potentially represent different clusters. VAT and iVAT seem to work well for relatively small size datasets. However, both have space and time complexity of $O(N^2)$, which limits their usefulness for input matrix sizes of an order of $10^5$ and so.  To overcome this limitation, an intelligent sampling based scalable clustering algorithm, clusiVAT~\cite{kumar2016hybrid} was proposed for visual cluster tendency assessment and subsequent clustering on big data. The essential steps in clusiVAT are:

\textbf{Maximin Random Sampling (MMRS):} MMRS sampling begins by finding $k'$ (an overestimate of $k$) Maximin samples (\textit{distinguished objects}), which are furthest from each other in the input data. Then, each object in the input data is grouped with its nearest Maximin sample with NPR. This step divides the entire data into $k'$ groups. Then, a sample $S$ of size $n$ (just a small fraction of input data size, $N$) is built by selecting a proportional number of random data points (Random sampling (RS)) from each of the $k'$ groups. Hence the term MMRS is used for the overall process. Any value of $k'$ which overestimates  assumed true number of clusters (k) i.e.,  $k'\geq k$ should be a good choice~\cite{hathaway2006scalable}. For $n$, $n$ = \textit{a few hundred} datapoints is a reasonable choice for most datasets~\cite{rathore2018approximate}.

\textbf{Step 2. iVAT:} The iVAT is applied to the small $D_{n}$ (computed from $n<<N$ samples) distance matrix to obtain its reordered distance matrix ${D'}_{n}^{*}$. The image $I({D'}_{n}^{*})$ usually provides an useful estimate of $k$, without the need to compute the very large distance matrix $D_{N}$.

\textbf{Step 3. Clustering:} Single linkage clusters are always aligned partitions in the VAT/iVAT reordered matrices. Having the estimate of $k$ from the previous step, we cut the $k-1$ longest edges in the iVAT-built MST of $D_{n}$, resulting in $k$ single linkage clusters. If the dataset is complex and clusters are intermixed, cutting the $k-1$ longest edges may not always be a good strategy as the outliers, which are typically furthest from normal clusters, might comprise most of the $k-1$ longest edges of the MST, resulting in misleading partitions. A useful approach in such a scenario is to manually select the dark blocks, and find the sample trajectories representing each dark block. Another useful approach~\cite{kumar2017visual}  to obtain clusters is by cutting the MST using cut threshold magnitudes ordered by edge distances $d$ in the MST. The cluster boundaries are defined by those indices $z$, which satisfy
\begin{align}
d_{z}> \alpha \times mean(d),
\end{align}
where $\alpha$ is a parameter that controls how far two groups of data points should be from each other to be considered as separate clusters. Smaller values of $\alpha$ represent tighter cluster boundaries, while large values of $\alpha$ create loose cluster boundaries. The procedure to find an optimal value of $\alpha$ is described in~\cite{kumar2017visual}.

\textbf{Step 4. Nearest Prototyping Rule (NPR):} The $k$-partitions of the $n$ samples are non-iteratively extended to the remaining (non-sampled) $N-n$ objects in the dataset using the nearest prototyping rule.

The implementation and pseudocodes of the trajDTW, VAT, iVAT, and clusiVAT algorithms are well documented in the literature, and are not produced here for brevity.
\subsection{Markov Chain Model}\label{markovchainmodel}
A \textit{Markov chain} (MC) is the simplest form of the Markov process in which only the current state determines the probability of transitioning to the next state. Specifically, a Markov chain model is defined by the transition matrix $M$, which contains the transition probabilities associated with various state changes. In a road network, an MC is constructed by assigning a state to each node or road segments in the given road network. For any two adjacent road segments $R_{i}$ and $R_{j}$ in road network $G_{RN}$, the transition probability of traveling from $R_{i}$ to $R_{j}$ in one step is given by
\begin{align}\label{Eq4}
p_{ij}=\frac{\#(R_i,R_j)}{\#(R_i)},
\end{align}
where $\#(R_i,R_j)$ is the number of trajectories that contain the sequence $\{R_i,R_j\}$ and $\#(R_i)$ is the total number of trajectories that passes through $R_i$. For each pair of adjacent road segments in the graph network, the transition probabilities can be computed using~(\ref{Eq4}), and stored as entries $M_{ij}$ of transition probability matrix $M$ (of size $|E| \times |E|$). We also define a transition count matrix $W$  whose $ij$-th entry $W_{ij}$ represents the number of trajectories that contain sequence $\{R_i,R_j\}$ i.e., $W_{ij} = \#(R_i,R_j)$. We utilize $W$ in computing a representative trajectory for each cluster in our work.

\setlength\belowcaptionskip{-2ex}

\begin{figure*}
\centering
\includegraphics[width=0.95\textwidth]{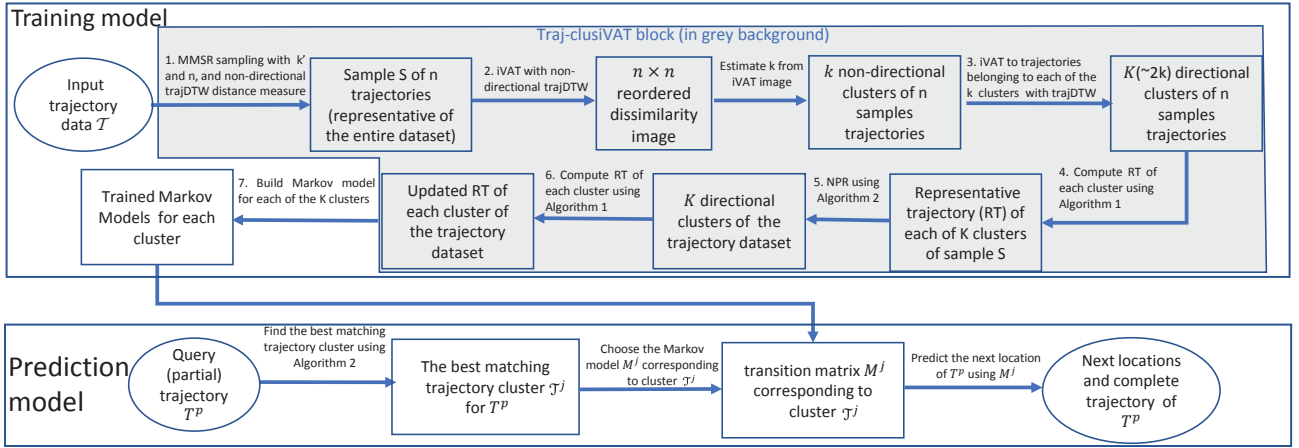}
\caption{The architecture of our proposed framework.}
\label{Fig:TrajPredFramework}
\end{figure*}

\section{Proposed Framework}\label{proposedframework}
This section presents our proposed framework for trajectory prediction. The frequent route patterns of moving objects can be discovered by clustering their historical trajectories. In our framework, we employ a modified version of the clusiVAT algorithm that we called Traj-clusiVAT. In Traj-clusiVAT, we introduce a \textit{representative trajectory} for each cluster to improve the performance of NPR for trajectory clustering. We also modify the NPR technique in Traj-clusiVAT to improve its performance for trajectory prediction. The Traj-clusiVAT algorithm partitions the trajectories into different groups of similar trajectories, based on the trajDTW distance measure. After clustering trajectories, we train a first-order Markov chain model for each cluster using only the trajectories contained therein. Then, these trained Markov chain models are used for trajectory prediction. The architecture of our proposed framework consisting of both training and prediction models is illustrated in~\Fig~\ref{Fig:TrajPredFramework}.

\subsection{Training Model}
The essential steps of our training model are: (i) MMRS sampling on input trajectory data, (ii) VAT/iVAT and clustering the trajectory sample using non-directional trajDTW to obtain  $k$ non-directional clusters (iii) VAT/iVAT and clustering the trajectories of each of the $k$ clusters using trajDTW resulting in $K$ (approx. $2k$) directional clusters (iv) Compute \textit{representative trajectory} (RT) of each cluster, (v) Assign remaining non-sampled trajectories to $K$ clusters using NPR (vi) Re-compute the RT of each cluster, and (vii) Train a first-order Markov chain model for each cluster.  The first six steps constitute the Traj-clusiVAT clustering algorithm. Below, we explain each step corresponding to the steps as shown in~\Fig~\ref{Fig:TrajPredFramework}.

\subsubsection{MMRS sampling on input trajectory data}
The first step consists of extracting a small, representative sample from the large trajectory data using MMRS sampling with non-directional trajDTW distance measure on input trajectory data $\mathcal{T}$. The aim of this step is to find the most distinguished vehicle routes in a given road network. The use of non-directional trajDTW circumvents the selection of more than one trajectory from the same route. In this way, the Maximin (first) step of MMRS ensures that MMRS samples contain the $k'$ MM trajectories of the most distinguished vehicle routes. This divides the trajectory data $\mathcal{T}$ into $k'$ partitions. Then, additional trajectories are randomly chosen from each of the $k'$ partitions to generate a sample $S$ of $n$ trajectories. The MMRS intelligently chooses $n$ trajectories which are almost equally distributed among the different clusters as the $N$ trajectories in the big trajectory data, i.e., it obtains a representative sample.
\subsubsection{Clustering trajectory samples using non-directional trajDTW}
The previous step provides a trajectory sample $S$ containing $n$ trajectories. In this step, the iVAT is applied to the distance matrix $D_{n}$ returning a reordered distance matrix ${D'}_{n}^{*}$, and the cut magnitudes of the MST links, $d$. The visualization of ${D'}_{n}^{*}$ using $I({D'}_{n}^{*})$ suggests the number of clusters $k$ present in the dataset. The $k$ partitions can be obtained by cutting the $k-1$ edges or by cutting the MST cut magnitudes $d$ using cut threshold $\alpha$, as mentioned in Section~\ref{clusiVATAlgo}. The non-directional trajDTW distance measure is used in this step to cluster the $n$ trajectories in order to avoid incorrect clustering due to the movement direction of trajectories, as mentioned in Section~\ref{NonDirectTrajDTW}). From here on, we denote $k$ as the number of non-directional clusters.

\subsubsection{Clustering trajectories in each cluster using trajDTW}
The previous step clusters the trajectories based on their path similarity computed using non-directional trajDTW, which ensures that the trajectories that are in opposite directions, but follow similar routes, are clustered together. Since Markov chain models are used in our framework to model the trajectories of each cluster, their transition probabilities may be misleading for trajectory prediction task for clusters in which the number of trajectories in opposite directions is approximately equal. To circumvent this problem, we use the trajDTW (directional) distance measure for the sample trajectories of each cluster obtained in the previous step to separate the trajectories going in opposite directions using a second application of the iVAT algorithm, which in turn, gives $K \sim 2k$ directional clusters.
\subsubsection{Computing the RT of each cluster}
In the NPR (next) step of clusiVAT, the non-sampled trajectories are assigned to one of the clusters (found in the previous step) based on their (nearest) distances from each cluster. For a fast and reliable implementation of NPR, we require a \textit{representative trajectory }(RT) for each cluster that best describes the cluster, much like centroid-based clustering methods identify a representative "center" for each cluster. However, it is not possible to compute the centroid of trajectory clusters in a conventional way due to different lengths of trajectories in each cluster. Existing methods of calculating RT~\cite{yavacs2005data,lee2008traclass,lu2011mining,sung2012trajectory} in the literature either compute the mean trajectory using the average of GPS coordinates~\cite{lee2008traclass,sung2012trajectory}; or select a trajectory from each cluster which minimizes the dissimilarity between all the trajectories within the cluster~\cite{yavacs2005data,lu2011mining}; or pick a random trajectory~\cite{lu2011mining} from each cluster, and designates it as the RT. These methods incur a large computational cost to compute an RT that minimizes the dissimilarity among all the trajectories. Additionally, RTs computed using these methods do not show all the possible variability inside a cluster~\cite{ferreira2013vector}.  The mean trajectory computed from trajectories of different lengths may be inaccurate; thus, it may not be a good representative of the cluster.

Our scheme generates an \textit{imaginary trajectory} (IT) (it may not belong to any of the trajectories in the cluster) as an RT for each cluster that describes the major movement patterns of the trajectories belonging to that cluster. The pseudocode of our proposed method to compute RT for each cluster is shown in Algorithm~\ref{Algo1}. Below, we explain our RT computing algorithm.

First, we compute the transition count matrix $W^{i}$ for each cluster $\mathcal{T}^{i}$ using the trajectories in that cluster (line $2$). Then, for each cluster $\mathcal{T}^{i}$, we compute the set of \textit{frequent road segments} (FRSs) $\mathcal{R}_{FRS}^{i}$  using the $MinT$ threshold (line $3$). The road segments in cluster $\mathcal{T}^{i}$ which contains at least $MinT \%$ of the total trajectories in that cluster are assigned to $\mathcal{R}_{FRS}^{i}$. Then, a set of  \textit{frequent source segments} (FSSs) $\mathcal{R}_{FSS}^{i}$ is identified (line $4$). A source segment $R_{SS}^{i}$ is a FSS, $R_{FSS}^{i}$, if at least $MinT \%$ of total trajectories in the cluster originate from $R_{SS}^{i}$  i.e, $R_{FSS}^{i} \in \mathcal{R}_{SS}^{i},~R_{FSS}^{i} \in \mathcal{R}_{FRS}^{i}$. Then for each FSS $R_{FSS}^{i} \in \mathcal{R}_{FSS}^{i}$ (line $5$),  an imaginary trajectory $IT^{i}(R_{FSS}^{i})$ is initialized with $R_{FSS}^{i}$ assigning it as current segment $R_{current}$ (lines $6-7$). In lines $9-17$, we compute the next RS $R_{next}$ based on the highest transition count from current RS $R_{current}$ using transition count matrix $W^{i}$ (refer to Section~\ref{markovchainmodel}) . If $R_{next} \in \mathcal{R}_{FRS}^{i}$, then $R_{next}$ is added to current $IT^{i}(R_{FSS}^{i})$, and assigned as $R_{current}$ to compute new $R_{next}$. The steps in lines $9-18$ are repeated until $R_{next}$ is non-FRS, \textit{which means an imaginary trajectory is an ordered sequence of only frequent road segments in that cluster}. A total of $|\mathcal{R}_{FSS}^{i}|$ imaginary trajectories will be generated for each cluster $\mathcal{T}^{i}$, corresponding to each $R_{FSS}^{i} \in \mathcal{R}_{FSS}^{i}$. We define a variable $Count\_score$  (line $8$) for each imaginary trajectory  $IT^{i}(R_{FSS}^{i}),~R_{FSS}^{i} \in \mathcal{R}_{FSS}^{i}$,  which is the sum of the total transition counts of each RS $\in IT^{i}(R_{FSS}^{i})$ in cluster $\mathcal{T}^{i}$. Among all $|\mathcal{R}_{FSS}^{i}|$ ITs, the one which has the highest $Count\_score$ will be assigned as $RT(\mathcal{T}^{i})$ of cluster $\mathcal{T}^{i}$ (line $20$). As the $RT(\mathcal{T}^{i})$ is the sequence of FRS with highest $Count\_score$, it contains major movement behaviour or patterns of the trajectories belonging to the cluster $\mathcal{T}^{i}$.

Algorithm~\ref{Algo1} does not require the computation of dissimilarity among all trajectories in that cluster to compute RT, which is computationally expensive for large size clusters. In contrast, Algorithm~\ref{Algo1} is a novel algorithm to compute RT based on the transition count matrix of each cluster.

\begin{algorithm}
\caption{Computing the RT of each cluster}\label{Algo1}
\textcolor{black}{
\begin{algorithmic}
\State  \textbf{Input:} $\mathcal{T}^{j}$- set of trajectories in cluster $j$,  $N_{j}$- number of trajectories in cluster $j$, $\mathcal{R}^{j}$- set of road segments in cluster $j$, $\mathcal{C}(\mathcal{T})=\{\mathcal{T}^{1},...,\mathcal{T}^{K}\}$- set of cluster of trajectories, $MinT$- FRS threshold
\end{algorithmic}
\begin{algorithmic}[1]
\For{each cluster $\mathcal{T}^{i} \in \mathcal{C}(\mathcal{T})$}
\State Compute transition count matrix $W^{i}$ for cluster $\mathcal{T}^{i}$
\State Compute FRSs, $\mathcal{R}_{FRS}^{i}$, from $\mathcal{R}^{i}$,  $\mathcal{R}_{FRS}^{i}$ =  $\{R_{j} \in  \mathcal{R}^{i}\}_{\#(R_{j}) \geq  MinT \times N_{i}}$
\State Compute FSSs, $\mathcal{R}_{FSS}^{i}$ from $\mathcal{R}_{SS}^{i}$ = $\{R_{j}\}_{R_{j} \in \mathcal{R}_{SS}^{i} \in \mathcal{R}_{FRS}^{i}}$
\For{each FSS $R_{FSS}^{i} \in \mathcal{R}_{FSS}^{i}$}
\State Assign $R_{FSS}^{i}$ as current road segment, $R_{current} = R_{FSS}^{i}$
\State  Initialize an imaginary trajectory $IT$ with $R_{current}$,  $IT^{i}(R_{FSS}^{i}) = \{R_{current}\} $
\State $Count\_score(IT^{i}(R_{FSS}^{i}))=0$
\While {each RS of $IT^{i}(R_{FSS}^{i})  \in \mathcal{R}_{FRS}^{i}$}
\State Compute next RS, $R_{next} = \displaystyle \operatorname*{arg\,max}_{R_{j} \in \mathcal{R}^{i}}~\{W_{current,j}^{i} \}$
\If {$R_{next} \in \mathcal{R}_{FRS}^{i}$}
\State  Append $R_{next}$ to existing $IT^{i}(R_{FSS}^{i})$
\State $R_{current}= R_{next}$
\State $Count\_score(IT^{i}(R_{FSS}^{i}))$ += $W_{current,next}^{i}$
\Else
\State \textbf{break;}
\EndIf
\EndWhile
\EndFor
\State Select $IT^{i}(R_{FSS}^{i})$ with the highest $Count\_score(IT^{i}(R_{FSS}^{i}))$ from all $|\mathcal{R}_{FSS}^{i}|$ $IT$s of $\mathcal{T}^{i}$, and assign it as RT for cluster $\mathcal{T}^{i}$
\EndFor
\end{algorithmic}
\begin{algorithmic}
\State  \textbf{Output:} $RT(\mathcal{T}^{i})$- RT for each cluster $\mathcal{T}^{i} \in \mathcal{C}(\mathcal{T})$
\end{algorithmic}}
\end{algorithm}

\begin{algorithm}
\caption{Hybrid NPR Method}\label{Algo2}
\textcolor{black}{
\begin{algorithmic}
\State  \textbf{Input:} $T_{q}$ - query trajectory, $M^{j}$ - transition probability matrix for cluster $j$,  $RT(\mathcal{T}^{j})$- representative trajectory for cluster $j$
\end{algorithmic}
\begin{algorithmic}[1]
\State Compute the path probability $P^{i}(T_{q})$ of query trajectory in each cluster $\mathcal{T}^{i}$ using $M^{i}$ and Eq.~\ref{Eq.5}.
\If {any($P^{i}(T_{q}) > 0$} \Comment{if $T_{q}$ is present in any cluster $\mathcal{T}^{i}$ }
\State Select the cluster $c$ with the highest $P^{i}(T_{q})$ i.e., $c = \displaystyle \operatorname*{arg\,max}_{\mathcal{T}^{i} \in \mathcal{C}(\mathcal{T})}~ \{P^{i}(T_{q})\}$
\Else
\State Compute the trajDTW distance of $T_{q}$ from $RT(\mathcal{T}^{i})$, $y^{i} = trajDTW(T_{q},RT(\mathcal{T}^{i}))$, for each cluster $\mathcal{T}^{i} \in \mathcal{C}(\mathcal{T})$
\State Select the cluster $c$ with the minimum $y^{i}$ i.e., $c= \displaystyle \operatorname*{arg\,min}_{\mathcal{T}^{i} \in \mathcal{C}(\mathcal{T})}~ \{y^{i}\}$
\EndIf
\State Assign the $T_{q}$ with cluster $c$ (or $\mathcal{T}^{c}$).
\end{algorithmic}
\begin{algorithmic}
\State  \textbf{Output:} cluster label for $T_{q}$
\end{algorithmic}}
\end{algorithm}

\begin{figure*}
\centering
\includegraphics[width=0.95\textwidth]{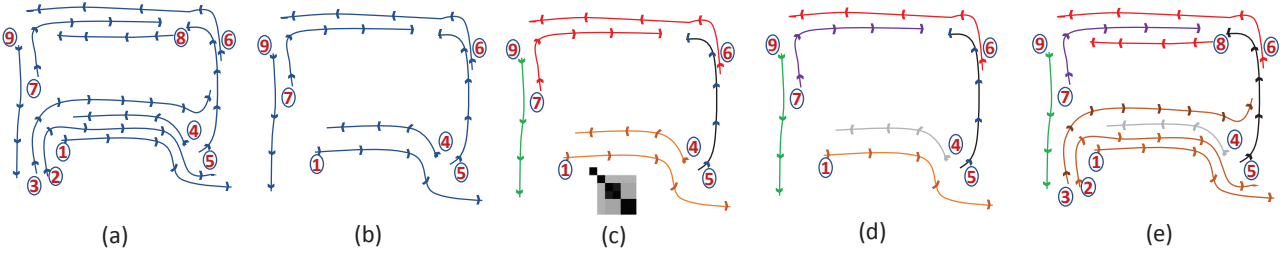}
\caption{A simple illustration of Traj-clusiVAT for trajectory clustering}
\label{Fig:clusiVATillustration}
\end{figure*}

\subsubsection{Assigning non-sampled trajectories to identified $K$ clusters using NPR}
The previous step gives representative trajectory $RT(\mathcal{T}^{i})$ for each cluster $\mathcal{T}^{i}$. In this step, $N-n$ non-sampled trajectories are assigned to one of the $K$ directional clusters based on the NPR. The NPR method in clusiVAT uses the trajDTW (directional) distance measure to assign non-sampled trajectories to one of the $K$ clusters based on their nearest distance from (clustered) sample trajectories. However, trajDTW distance of a non-sampled trajectory to cluster RTs may not be an appropriate measure for the NPR step due to its dependency on the \textit{length} of trajectories, as explained by the following example.

Suppose $T_{a}$ is a non-sampled trajectory in $\mathcal{T}$, and $RT(\mathcal{T}^{i})$ and $RT(\mathcal{T}^{j})$ are the RT of cluster $\mathcal{T}^{i}$ and $\mathcal{T}^{j}$, respectively.  Let $T_{a}$ be a sub-trajectory of  $RT(\mathcal{T}^{i})$  i.e., $T^{a}$ is fully contained in $RT(\mathcal{T}^{i})$. Since the trajDTW distance   relies on a warping window size parameter $w$, the $trajDTW(T_{a},RT(\mathcal{T}^{i}))$  not only depends on the coordinates of RSs of both trajectories, but it also depends on the length of $T_{a}$ and $RT(\mathcal{T}^{i})$. Moreover,  $trajDTW(T_{a},RT(\mathcal{T}^{i}))$ also varies depending on the position of $T_{a}$ in $RT(\mathcal{T}^{i})$ due to window parameter.
Therefore, even if $T_{a} \sqsubseteq RT(\mathcal{T}^{i})$ and  $T_{a} \not \	RT(\mathcal{T}^{j})$, $T_{a}$ may be incorrectly assigned to cluster $\mathcal{T}^{j}$ instead of $\mathcal{T}^{i}$ if  $trajDTW(T_{a},RT(\mathcal{T}^{i})) \geq  trajDTW(T_{a},RT(\mathcal{T}^{j}))$. Here is such an example from T-Drive data. Suppose $T_{1} = \langle 70,75,90,89,88  \rangle$  is a non-sample trajectory, and $RT(\mathcal{T}^{1}) = \langle 16,18,68,70,75,90,89,88 \rangle$ and $RT(\mathcal{T}^{2}) = \langle 68,70,75,91,92 \rangle$ are RTs of two clusters, where each trajectory is represented by a sequence of road segments' IDs of Beijing road network (Refer to Section~\ref{Dataset}). The trajDTW distances are: $trajDTW(T_{1},RT(\mathcal{T}^{1})) = 0.3482$ and $trajDTW(T_{1},RT(\mathcal{T}^{2})) = 0.2767$. Therefore, although  $T_{1}$ is  a sub-trajectory of  $RT(\mathcal{T}^{1})$, it will be assigned to cluster $\mathcal{T}^{2}$ based on nearest trajDTW distance. Such assignments of non-sampled trajectories to (incorrect) cluster may include outlier trajectories or road segments in that cluster, which may adversely affect Markov chain modeling, and consequently, degrade the performance of trajectory prediction.

To address above issue, we propose a \textit{hybrid} NPR strategy based on the path probability and trajDTW distance measure.  \textit{Hybrid} NPR is similar to clusiVAT NPR except for those non-sampled trajectories, which are sub-trajectory of any of the clusters' trajectories. The pseudocode of our hybrid NPR method is shown in Algorithm~\ref{Algo2}. For a query trajectory $T^{q}= \{R_{1},R_{2},...,R_{l}\}$, we first compute the path probability $P^{i}(T^{q})$ for each cluster $\mathcal{T}^{i}$, which is defined as
\begin{align}\label{Eq.5}
P^{i}(T^{q})= \prod_{j=1}^{l} p_{j(j+1)} \Leftrightarrow  \prod_{j=1}^{l} M_{j(j+1)}^{i}.
\end{align}
$P^{i}(T^{q}) > 0$ means that sequence $T^{q}$ appears at least once in cluster $\mathcal{T}^{i}$, whereas $P^{i}(T^{q}) = 0$ means that sequence $T^{q}$ is not present in cluster $\mathcal{T}^{i}$. If the sequence $T^{q}$ is present in any cluster $\mathcal{T}^{i}$ i.e., any($P^{i}(T^{q})) > 0$, then  $T^{q}$ is assigned to the cluster with the highest path probability. If the sequence $T^{q}$ is not present in all clusters $\mathcal{T}^{i}$ i.e., all($P^{i}(T^{q})) = 0$, then  $T^{q}$ is assigned to the cluster based on its (minimum) trajDTW distance from RTs. All non-sampled trajectories in $\mathcal{T}$ are assigned to one of the $K$ clusters using Algorithm~\ref{Algo2}.

\subsubsection{Recompute the RT of each cluster after NPR}
The assignment of all non-sampled trajectories to one of the $K$ clusters in the NPR step updates each cluster with new trajectories. Therefore, a representative trajectory is recomputed for each updated cluster using Algorithm~\ref{Algo1}.
\subsubsection{Train Markov chain model}
For each of the $K$ clusters, we build a first-order Markov chain model using the trajectories of that cluster. Specifically, we compute the transition probability matrix $M^{i}$ for each cluster $\mathcal{T}^{c}$.

For a basic understanding of  Traj-clusiVAT algorithm, we graphically explain its steps on a small trajectory data $\mathcal{T}$, as shown in Fig~\ref{Fig:clusiVATillustration}. An input trajectory data $\mathcal{T}$ containing $N=9$ trajectories is shown in Fig~\ref{Fig:clusiVATillustration} (a). The MMRS sampling on $\mathcal{T}$ with non-directional trajDTW in the first step returns a MMRS sample $S$ containing $n=6$ sample trajectories $\{1,4,5,6,7,9\}$, which are well-distributed in sample $S$, as shown in Fig~\ref{Fig:clusiVATillustration} (b). In the next step, iVAT is applied to $S$ using the non-directional trajDTW distance measure, which clusters the trajectories based on the path similarity irrespective of their movement directions. The iVAT image in Fig~\ref{Fig:clusiVATillustration} (c) shows four dark blocks along its diagonal, which indicates four clusters in sample $S$. Having an estimate of $k=4$, sample $S$ is partitioned into four (non-directional) clusters $\{\{1,4\}, \{5\}, \{6,7\}, \{9\}\}$, as  shown with four different colors in Fig~\ref{Fig:clusiVATillustration} (c). Then, the trajectories in each cluster going in opposite directions are separated using the iVAT with the trajDTW  distance measure, which gives $K=6$ directional clusters $\{\{1\},\{4\},\{5\}, \{6\}, \{7\}, \{9\}\}$, each cluster is shown with a different colour in Fig~\ref{Fig:clusiVATillustration} (d). Since there is only one trajectory in each cluster in this case, they are  the RTs for corresponding clusters. In the next step,  non-sampled trajectories $\{2,3,8\}$ are assigned to one of the $6$ clusters using NPR (Algorithm~\ref{Algo2}), which partitions the complete data into $6$ clusters $\{\{1,2,3\},\{4\},\{5\}, \{6,8\}, \{7\}, \{9\}\}$. Trajectory $4$ is in different cluster than $\{1,2,3\}$ due to opposite direction. Then, a Markov chain model is trained for each cluster using the trajectories of that cluster.
\subsection{Prediction Model}
For a given partial trajectory $T^{p}= \langle L_{1},L_{2},...,L_{m} \rangle $, we first estimate the best  matching representative cluster $\mathcal{T}^{c}$ using our hybrid NPR approach, and then choose the corresponding Markov model of the cluster to predict the next locations $L_{i}$, $i\geq m+1$. Using the cluster $\mathcal{T}^{c}$, the location $L_{m+1}$ that the object will arrive at next is given by
\begin{align}
L_{m+1} = \displaystyle \operatorname*{arg\,max}_{L_{j} \in \mathcal{R}^{c}}~ \{p_{mj}\} \Leftrightarrow  \displaystyle \operatorname*{arg\,max}_{L_{j} \in \mathcal{R}^{c}}~ \{M^{c}_{mj}\}
\end{align}
The $T^{p}$ is updated with the next predicted location $L_{m+1}$. Then, the updated $T^{p}$  is used to estimate the best matching cluster and the corresponding MM is used to predict the next location. The complete trajectory is predicted by computing next locations in a sequential manner using these steps.
\section{Time Complexity}\label{sec:computationalcomplexity}
In this section, we discuss the time complexity of our proposed Traj-clusiVAT based TP approach.  The first step in Traj-clusiVAT is the selection of $k'$ distinguished trajectories which are at maximum distance from each other. This step has the time complexity of $O(k'N)$, where $k'$ is a user-defined parameter for an overestimate of the number of clusters in the input trajectory data and is usually chosen to be (inessentially) large (usually $50$ to $200$). The next step is to randomly select $n$ trajectories from $k'$ NPR groups to get a sample $S$. The computation of distance matrix $D_{n}$ and VAT on a sample $S$ has a time complexity of $O(n^2)$. Usually $n<<N$, so the computation of $D_{n}$ and VAT on $S$ is pretty fast and takes just a small fraction of the total run time of Traj-clusiVAT. The trajDTW distance measure uses Dijkstra's shortest path distance in the standard DTW algorithm. Its best, average case complexity with binary heaps is $O(|E| + |V| \log |V|)$~\cite{barbehenn1998note}. For two trajectories of length $l_{1}$ and $l_{2}$, standard DTW has time complexity of $O(l_{1}l_{2})$. Remark- There are approximate algorithms such as FastDTW~\cite{salvador2007toward} which have a linear time complexity in the average length of trajectories, however, we have not used this implementation in our experiments. The NPR step in Traj-clusiVAT has complexity of $O(n(N-n))$.  The computation of RTs has linear time complexity in $K$. The construction of a Markov model for each cluster is a simple and fast process, which has $O(K)$ time complexity and $O(|E|^2$) space complexity.

\section{Experiments}\label{sec:experiments}
In this section, we conduct an extensive experimental study on two real-life, vehicle trajectory datasets to evaluate the performance of our proposed framework. We first describe the datasets, their preprocessing, evaluation metrics and computational protocols adopted in our empirical study, and then present the experimental results.

\subsection{Datasets} \label{Dataset}
We performed our experiments on two real trajectory datasets.
\subsubsection{T-Drive taxi trajectory data~\cite{yuan2010t}} This trajectory dataset is obtained from the T-Drive project which contains one-week trajectories of $10,357$ taxis during the period of Feb. $2$ to Feb $8$, $2008$ within Beijing, China. The total number of points is about $15$ million and the total distance of the trajectories is $9$ million kilometres. In our experiment, we have taken a subset of this dataset, which contains trajectories from a road network in the center of   Beijing city, as shown in~\Fig~\ref{Fig:RoadNetworks}(a). This road network consists of $100$ nodes and $141$ road segments (edges). The average sampling interval is $177$ seconds with an average distance of about $623$ meters, which is quite large for a city traffic environment as the length of many road segments is smaller than the average sampling distance.
\subsubsection{Singapore taxi trajectory data} 
\begin{figure}
\centering
\subfloat[T-Drive: Road network in the center of Beijing]{\includegraphics[width=0.23\textwidth,height=0.1\textheight]{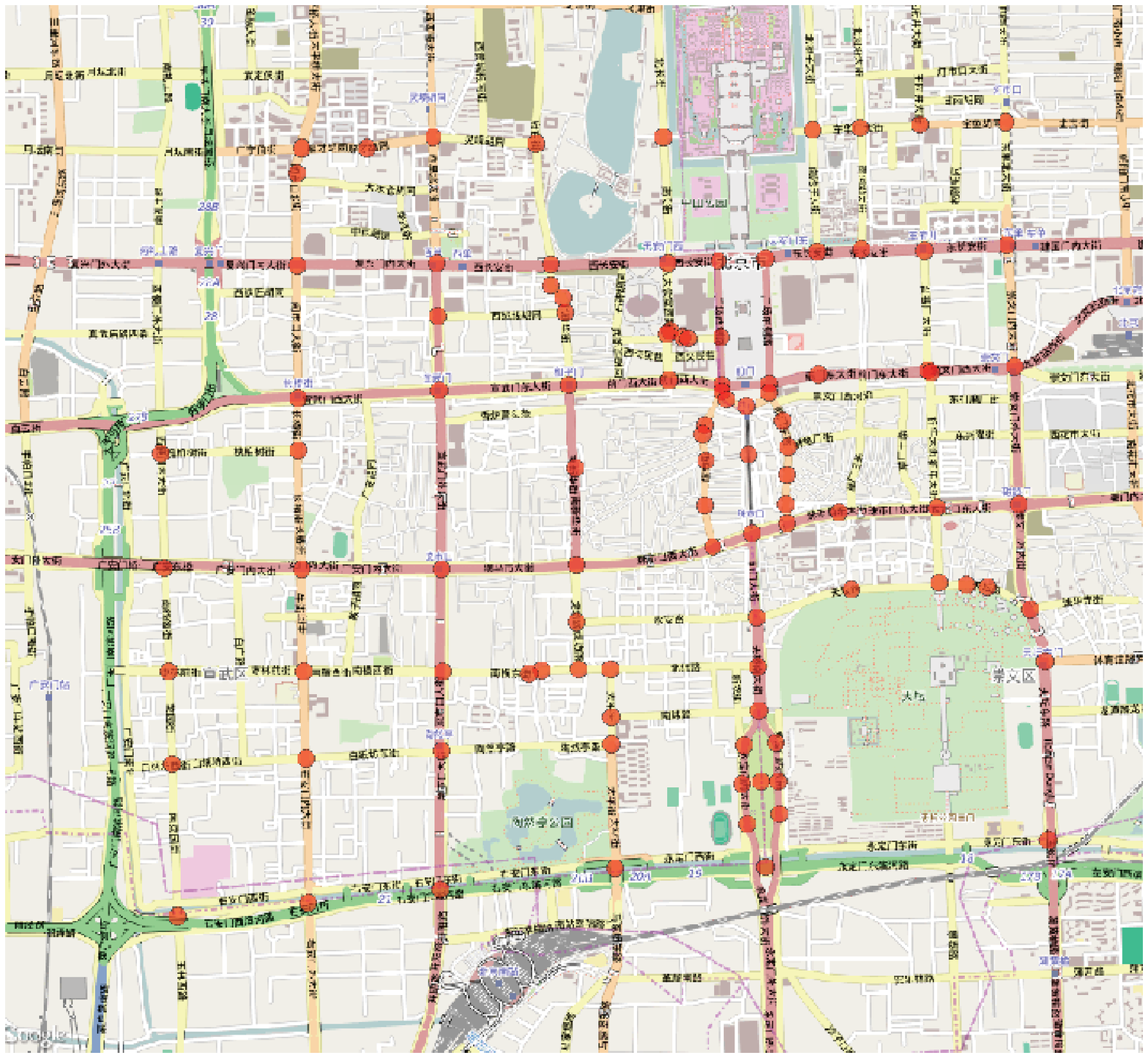}} \hfill
\subfloat[Singapore road network]{\includegraphics[width=0.25\textwidth]{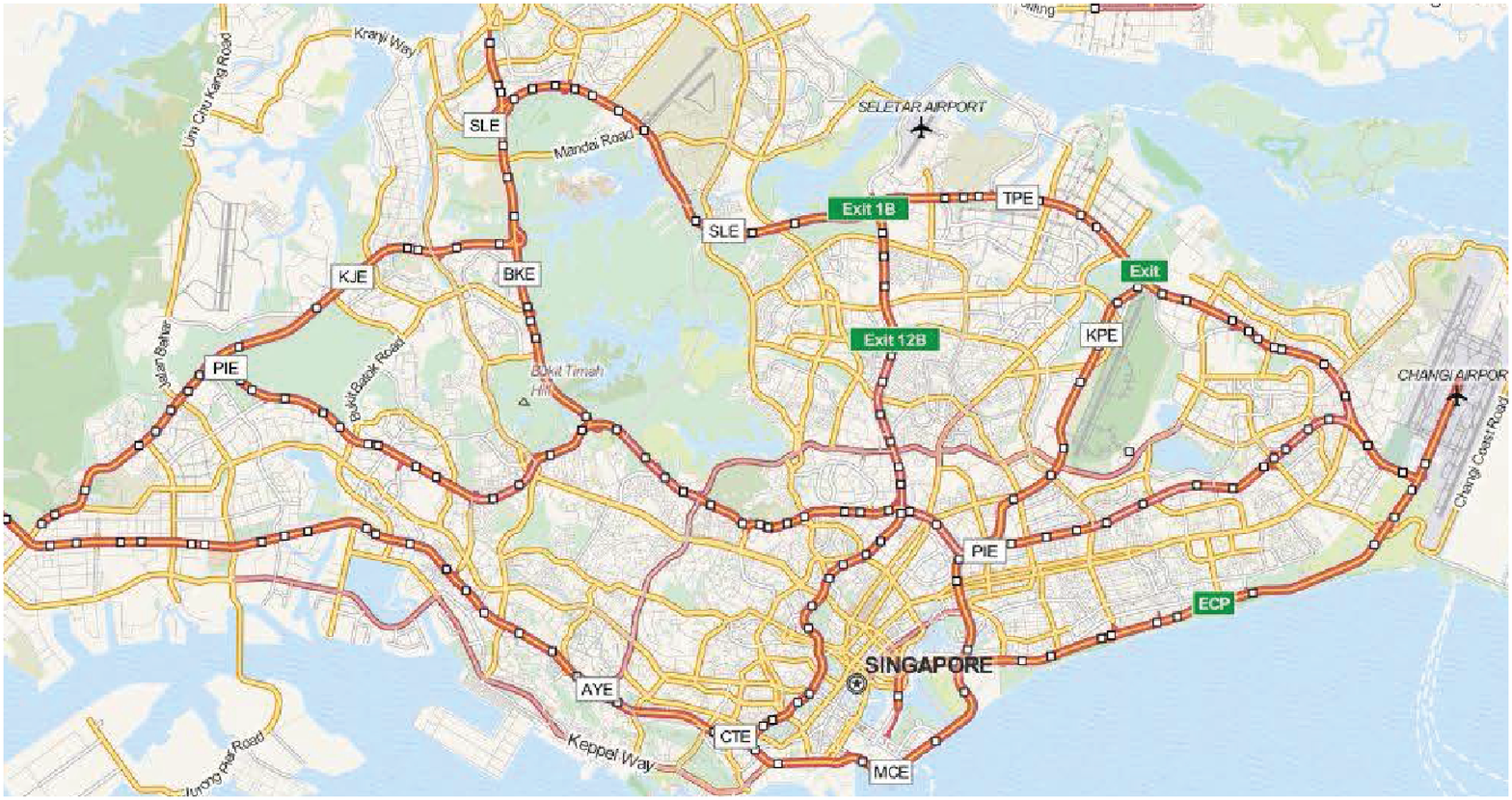}}
\caption{Road networks used in our trajectory prediction experiments}
\label{Fig:RoadNetworks}
\end{figure}

This dataset consists of the trajectories of more than $15,000$ taxis collected over a duration of $1$ month from a road network in Singapore City, as shown in~\Fig~\ref{Fig:RoadNetworks}(b). This dataset is very dense as it consists of more than $370$ million GPS logs. The general format of each data point is as follows: \{Time Stamp, Taxi Registration, Latitude, Longitude, Speed, Status\}. The Status field contains information about the occupation state of Taxi, such as \textit{FREE} and \textit{POB} (Passenger on Board). In order to extract each individual taxi's trip from the raw data, we detect the following sequence: starting from \textit{FREE} to \textit{POB} and ending from \textit{POB} to \textit{FREE}, using the trip extraction framework presented in~\cite{lu2015taxi}. This road network consists of $1641$ nodes and $2941$ edges, with an average edge length of $350$m.

\setlength\belowcaptionskip{-1ex}

\subsection*{Data Pre-processing}
To obtain the trajectories as a sequence of road segments, we use the popular open source map matching tool GraphHopper~\cite{GraphHopper}, which provides an implementation of the approach presented in~\cite{newson2009hidden}.

\begin{figure}
\centering
\subfloat[T-Drive Taxi]{\includegraphics[width=0.25\textwidth]{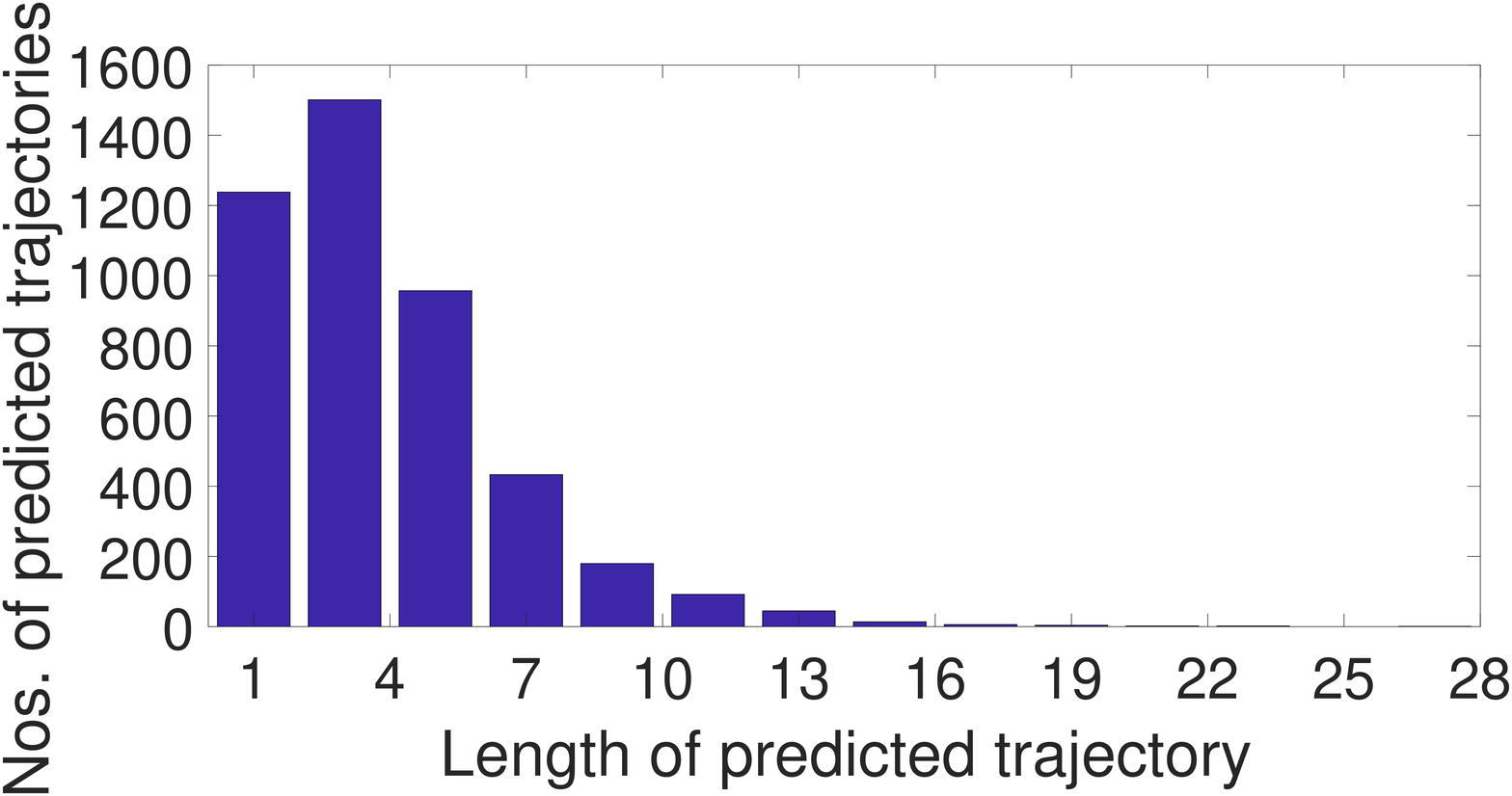}}
\subfloat[Singapore Taxi]{\includegraphics[width=0.25\textwidth]{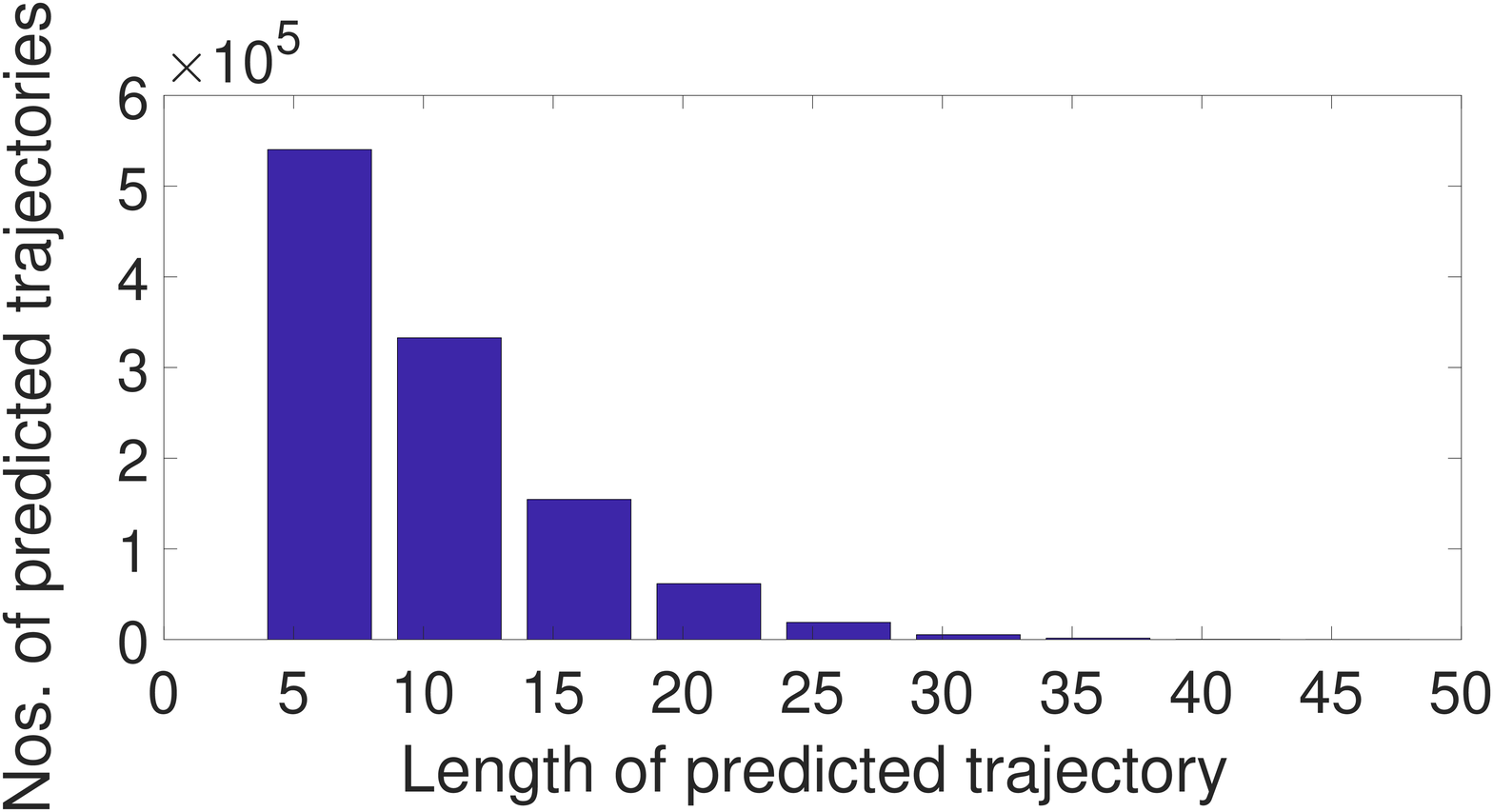}}
\caption{Trajectory distribution of predicted trajectories based on their lengths.}
\label{Fig:PredictedTrajectoryDistribution}
\end{figure}

After pre-processing, we have $N=43,405$ trajectories in the T-Drive data whose lengths lie in the range of $5$ to $200$ road segments and have an average of $14$ road segments, and $N=3,259,290$ ($3.26$ million) trajectories in the Singapore data whose lengths lie in the range of $10$ to $250$ road segments and have an average of $22$ road segments. To prepare training and test sets for both datasets, we first divided the trajectories into two sets based on the day of week viz., weekdays and weekends, during which the trip is being made. For the one-week T-Drive data, we considered trajectories during first $4$ weekdays  (Monday to Thursday) and first weekend day (Saturday) as the training set, and trajectories during the remaining days (Friday and Sunday) of that week as the test set. For the one-month Singapore data, we considered $60\%$  trajectories randomly as training set and remaining $40\%$ as the test set, for both weekdays and weekend data. The size of training and test sets for both trajectory datasets is shown in Table~\ref{Table2}.
We split each trajectory in a test set into two halves. The first half is used as a partial trajectory (or query trajectory) for predicting its future locations and the second half is used as ground truth to validate our predictions. The distribution of predicted trajectories (second half) in the T-Drive and Singapore test sets is shown in~\Fig~\ref{Fig:PredictedTrajectoryDistribution}.

\begin{table}[]
\centering
\caption{Training and test set description}
\label{Table2}
\begin{tabular}{|l|l|l|}
\hline
 & T-Drive Taxi & Singapore Taxi  \\ \hline
Training Set & $35,501$ & $1,955,573$ \\ \hline
Test Set & $7,904$ & $1,303,717$ \\ \hline
Total trajectories & $43,405$ & $3,259,290$ \\ \hline
\end{tabular}
\end{table}

\subsection{Evaluation Metrics}
In our experiments, we assess the performance of our framework for next location prediction (also known as one-step prediction) and long-route prediction using following evaluation metrics:
\subsubsection{Prediction Accuracy (PA)}
The PA is the ratio of correctly predicted locations to  the total possible number of predicted locations for each trajectory. Given a predicted trajectory sequence $T_{pred}=\{L_{1},L_{2},...,L_{m}\}$ and  a true (actual) trajectory sequence $T_{true}=\{R_{1},R_{2},...,R_{m}\}$,  the prediction accuracy is defined as
\begin{align}
PA=  \frac{1}{|T_{pred}|} \sum_{j=1}^{m} H(L_{j},R_{j}),
\end{align}
 where $H(L_{j},R_{j})$ is $1$ if $L_{j}=R_{j}$, else $0$. The average prediction accuracy is the average of PA for all predicted trajectories in test set $\mathcal{T}^{test}$.
\subsubsection{Prediction Rate (PR)} The PR is the number of trajectories that are correctly predicted over the total number of trajectories in test set. It is defined as
\begin{align}
PR=  \frac{1}{|\mathcal{T}^{test}|} \sum_{j=1}^{|\mathcal{T}^{tr}|} H(T_{pred_{j}},T_{true_{j}}),
\end{align}
where $H(T_{pred_{j}},T_{true_{j}})$ is $1$ if $T_{pred_{j}}=T_{true_{j}}$, else it is $0$.
\subsubsection{Distance error (DE)} Another important performance metric of the long-term prediction system is the capability of continuous route prediction. The \textit{distance error} is defined as the average spatial (\textit{Haversine}) distance between predicted and actual routes. Given a route sequence $T_{pred}$ and $T_{true}$, the distance error between them is given as
\begin{align}
DE(T_{pred},T_{true})=  \frac{1}{|T_{pred}|}\sum_{j=1}^{m} D_{H}(L_{j},R_{j}),
\end{align}
where  $D_{H}(L_{j},R_{j})$ is the Haversine~\cite{besse2017destination} distance between two locations (road segments).
\subsubsection{One-step accuracy (OA)} This is the ratio of correctly predicted next locations to the total predicted next locations for all trajectories in test set.
\subsubsection{One-step distance error (ODE)} The ODE defined as the average distance error for one-step (or next location) prediction.
\subsection{Comparison Methods}
Among the plethora of MM and clustering based TP methods available in the literature, we implemented these two approaches for comparison.
\begin{enumerate}
\item Mixed Markov model (MMM) based TP~\cite{asahara2011pedestrian}: MMM was proposed as an intermediate model between standard MM and HMM which can encompass all types of movement behaviour present in an input trajectory data. It first clusters the trajectories into groups using the EM algorithm, and then builds an MM for each group, which is subsequently used for prediction. This approach was tested on synthetic and real datasets in~\cite{asahara2011pedestrian}, which showed $74.1\%$ accuracy for MMM, in comparison to $16.9-45.6\%$ for MM and $2.4-4.2\%$ for HMM.
\item NETSCAN-based TP: The well-known density-based algorithm DBSCAN and its variants~\cite{huang2017mining,lei2011exploring,qiao2015traplan,jeung2007mining} have been used extensively as a trajectory clustering method for location prediction~\cite{jeung2008hybrid}. However, they are not suitable for a large number of trajectories as computation of the distance matrix is time intensive. Kharrat~\etal~\cite{kharrat2008clustering} proposed a trajectory clustering relative of DBSCAN, called NETSCAN which first finds dense road segments based on the moving object counts,  merges them to form dense paths on the road network, and then assigns sub-trajectories to the dense paths based on a measure of similarity. This method requires two user-defined parameters: a density threshold -- the minimal required density for transition, and a similarity threshold- the maximum density difference between neighbouring road segments.  We implement NETSCAN to cluster trajectories into dense road segments, then built an MM for each cluster, and subsequently used them for TP.
\end{enumerate}

\begin{figure*}
\centering
\subfloat[T-Drive Dataset]{\includegraphics[width=0.4\textwidth]{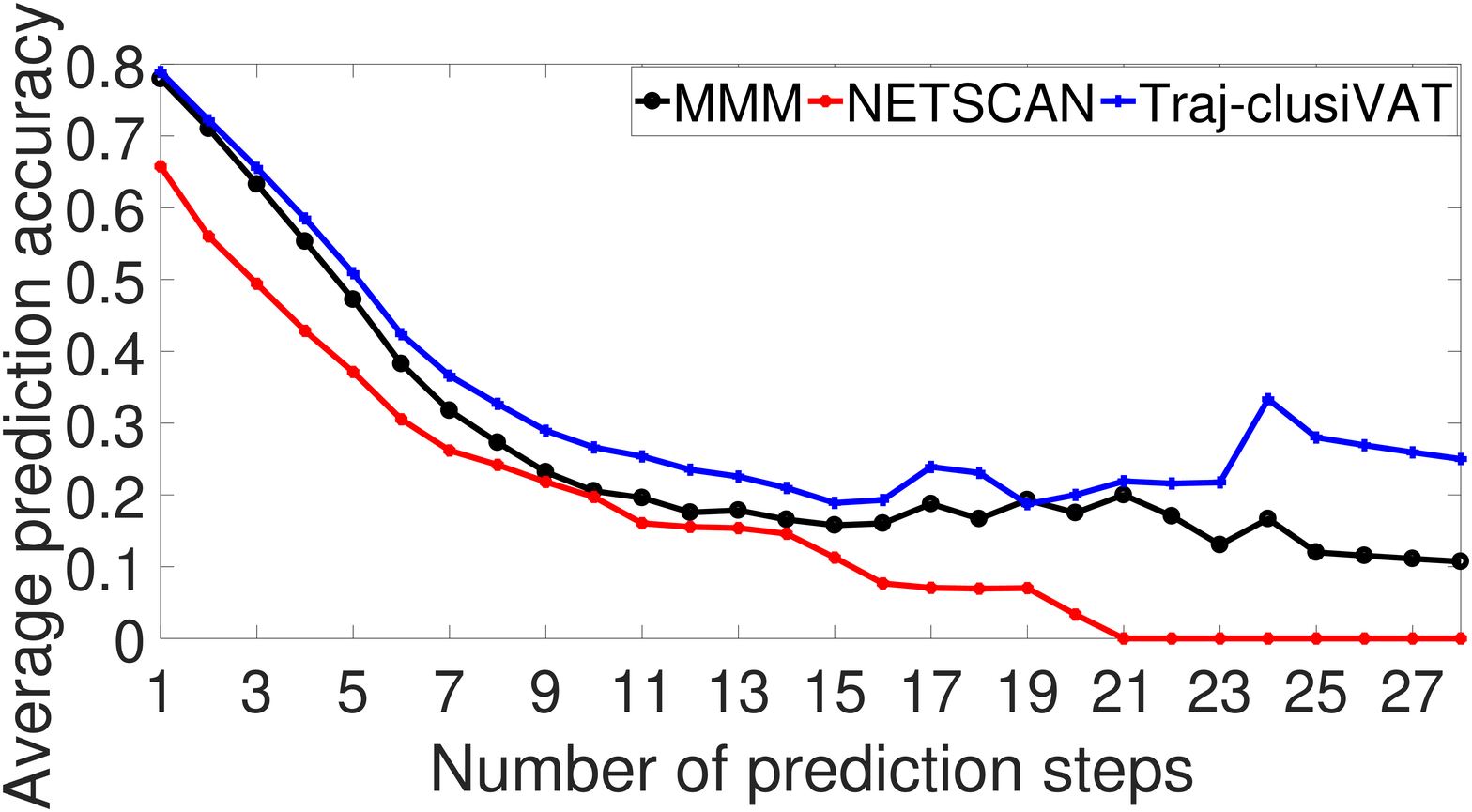}
\includegraphics[width=0.4\textwidth]{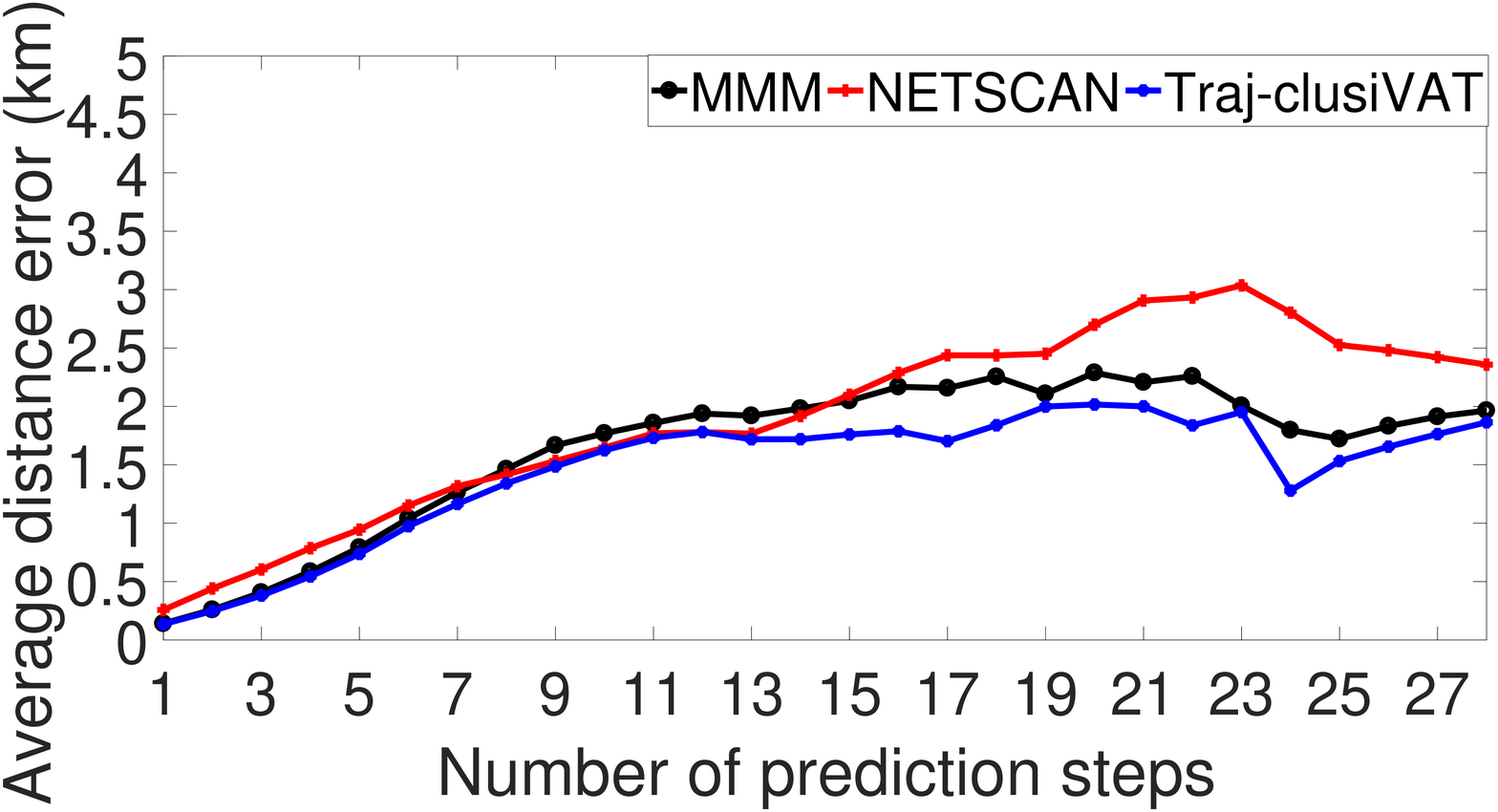}}\\
\subfloat[Singapore Taxi Dataset]{\includegraphics[width=0.4\textwidth]{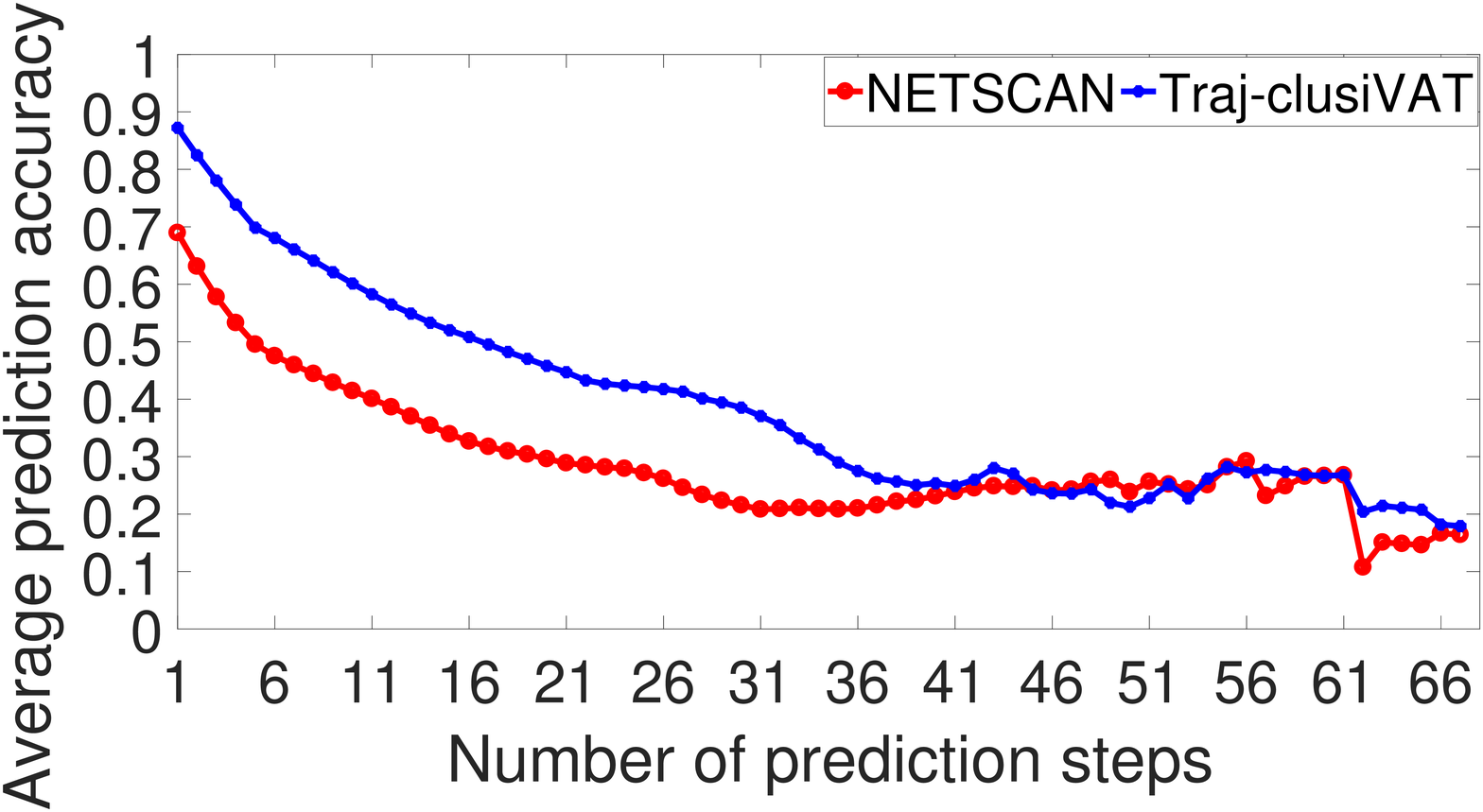}
\includegraphics[width=0.4\textwidth]{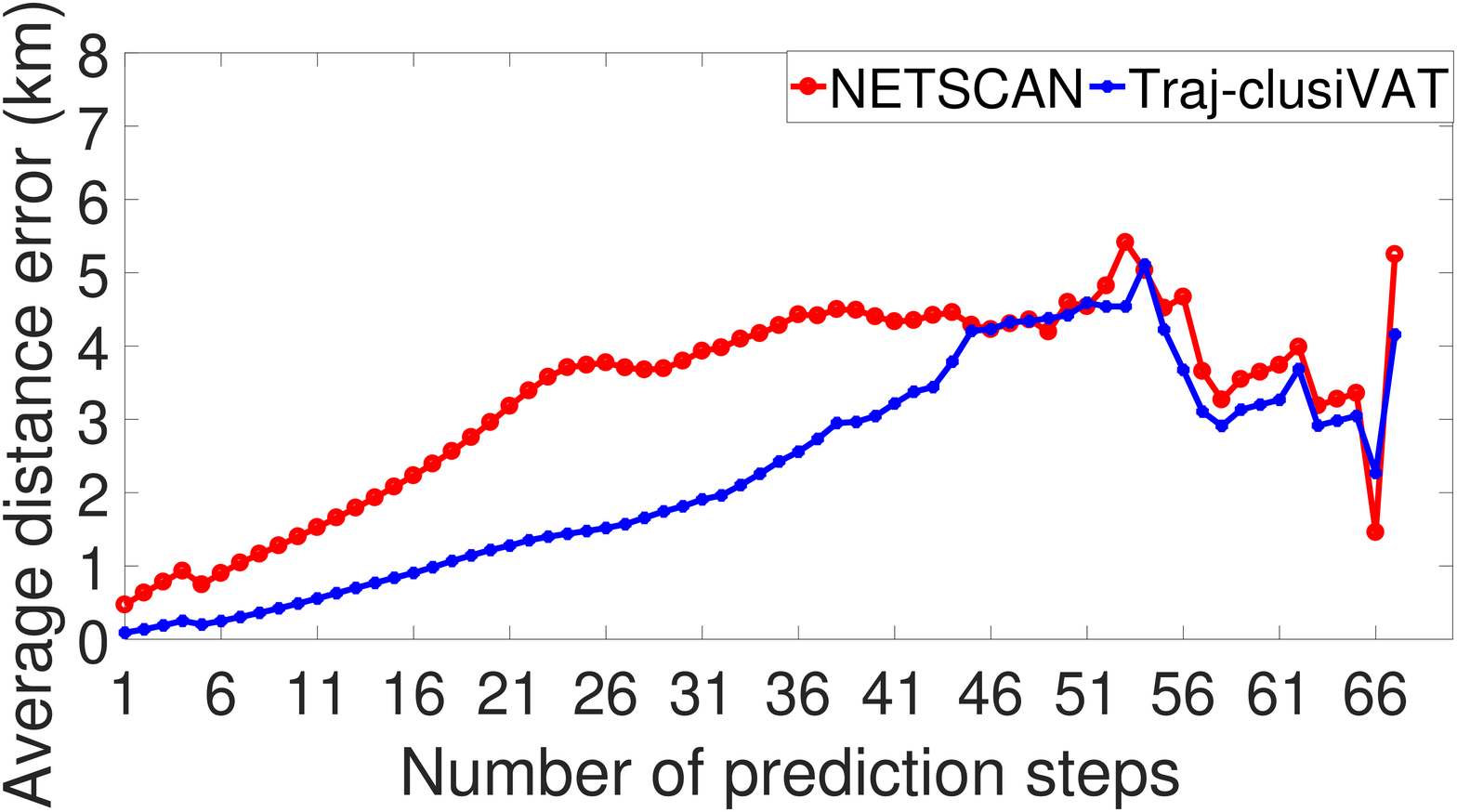}}\\
\caption{Average prediction accuracy and average distance error comparison by prediction steps}
\label{Fig:DEvsPredTrajLength}
\end{figure*}
Our proposed method and the baseline methods discussed above are also comparable in terms of prediction time  (which will be discussed shortly). They all require a short prediction time and satisfy the requirement of real-time prediction.
\subsection{Computation Protocols}
All algorithms were coded in MATLAB on a PC with the following configuration; OS: Windows $7$ ($64$ bit); processor: Intel Core $i7-4770$ $@3.40$GHz; RAM: $16$GB. We denote the comparison approaches of~\cite{asahara2011pedestrian} as MMM, of ~\cite{kharrat2008clustering} as NETSCAN, and our Traj-clusiVAT based TP approach as Traj-clusiVAT. All three algorithms  were applied to T-Drive data. The MMM method requires the computation and storage of an intermediate matrix of size $|E| \times |E| \times N$, which is very large for Singapore data (due to large $|E|$ and $N$), so we can not apply MMM to the Singapore data.  However, we have compared it with NETSCAN-TP and Traj-clusiVAT-TP on a subset obtained from a smaller part of the Singapore road network.   The number of mixed models of MMM was determined using $10$-fold cross-validation. The NETSCAN parameter, density threshold and similarity threshold, were chosen to get as many dense paths (with at least six road segments) as the number of clusters we get using the Traj-clusiVAT algorithm, for a fair comparison. The parameters for Traj-clusiVAT were chosen as follows: $k'=150$, $n=500$, and $\alpha =0.05$ for the T-drive data, and $k'=300$, $n=1000$, and $\alpha = 0.06$ for the Singapore data, and $MinT=30\%$ for both data. It is worth noting that, unlike other clustering algorithms, the clusiVAT algorithm is relatively insensitive to the choice of $k$ and $n$\cite{kumar2016hybrid}. Moreover, we study the effect of $\alpha$ on Traj-clusiVAT performance in our experiments.

\subsection{Comparison of MMM, NETSCAN, and Traj-clusiVAT for Long-term predictions}
Long-term prediction, also known as continuous route prediction, is a challenging and ongoing research problem in TP. In this experiment, we compare the performance of the MMM, NETSCAN, and Traj-clusiVAT-based prediction approaches for $m$-step predictions. Specifically, this refers to predicting the next $m$ locations for a given partial trajectory.~\Fig~\ref{Fig:DEvsPredTrajLength} shows the average prediction accuracy (left panels) and average distance error (right panels) of all three algorithms for increasing prediction steps. The graphs in~\Fig~\ref{Fig:DEvsPredTrajLength} support these observations:

(i) First, the Traj-clusiVAT outperforms the MMM and NETSCAN-based TP approaches based on the average PA and DE for the T-Drive data, as shown in~\Fig~\ref{Fig:DEvsPredTrajLength}(a). The higher the number of prediction steps, the larger the gap between Traj-clusiVAT and other two approaches. This means that the Traj-clusiVAT performs better not only for short-term predictions but it performs even better than the other two approaches for long-term predictions. This is probably because Maximin sampling in Traj-clusiVAT finds the trajectories which are furthest from each other. As the trajDTW distance measure yields higher distances for longer trajectories, Maximin sampling tends to pick longer trajectories in its output sample which form separate clusters in subsequent steps. The Markov models trained on these clusters after the NPR step contain all movement behaviours similar to those longer trajectory patterns. Therefore, if a query trajectory pattern is not available in any cluster, which is frequent for longer query patterns, then it is assigned to a cluster based on its nearest distance from all cluster RTs. This will assign longer query trajectories to any of the clusters containing longer trajectory patterns, and subsequently, corresponding MMs trained on these clusters contribute towards better predictions for longer query trajectories during the prediction phase. On the other hand,  the longer movement rules cannot be easily represented by Markov-based models, especially for irregular trajectory data, due to uncertainty in movement behaviours of vehicles in a complex road network. As there are only a few prediction trajectories available for the T-Drive test set whose lengths are greater than $16$ as shown in~\Fig~\ref{Fig:PredictedTrajectoryDistribution} (a), the performance of all approaches cannot be considered conclusive for longer prediction steps ($m>16)$  based on their performance on the T-drive data.

(ii) \Fig~\ref{Fig:DEvsPredTrajLength} (b) shows that the Traj-clusiVAT model also performs better than the NETSCAN-based method based on the average PA and DE values for the Singapore data. The gap between the NETSCAN and Traj-clusiVAT plots increases until $31$th prediction step and then reduces with longer prediction steps. This may be because the trajectory clusters obtained by NETSCAN are usually spread over the entire road network~\cite{Kumar2015ASF}, which results in longer dense paths. Therefore, its performance becomes competitive with Traj-clusiVAT for longer prediction lengths compared to its short-term prediction performance.

\textcolor{black}{(iii) It can be observed that difference in performance of the NETSCAN method and the proposed method is less for short-term and more for long-term prediction for T-drive data, whereas it is opposite for the Singapore data. This is because the T-drive subset contains many parallel and perpendicular road segments and intersections that span only a small road network (Beijing city center), whereas, Singapore data contains relatively longer and straight (fewer intersections) trajectories (compared to T-Drive subset) that span entire Singapore city road network (significantly bigger than Beijing city center road network). Therefore, incorrect predictions cause relatively smaller distance error for T-Drive data as compared to distance error for Singapore taxi dataset. NETSCAN's clusters for Singapore dataset are ordered sequences of only a few but longer and straight dense paths, therefore, it  modeled long trajectories better for Singapore dataset, and in turn, performed better for long-term prediction as compared to the T-Drive dataset.}

(iv) The performance of all three approaches deteriorates as the prediction step increases. This may be because the number of frequent trajectory patterns obtained is small for long-term predictions, which do not contain enough information to forecast future locations\footnote{And the other reason, as Niels Bohr said, is that "it is very hard to predict, especially the future"}.

\begin{figure*}
\centering
\subfloat[]{\includegraphics[width=0.25\textwidth]{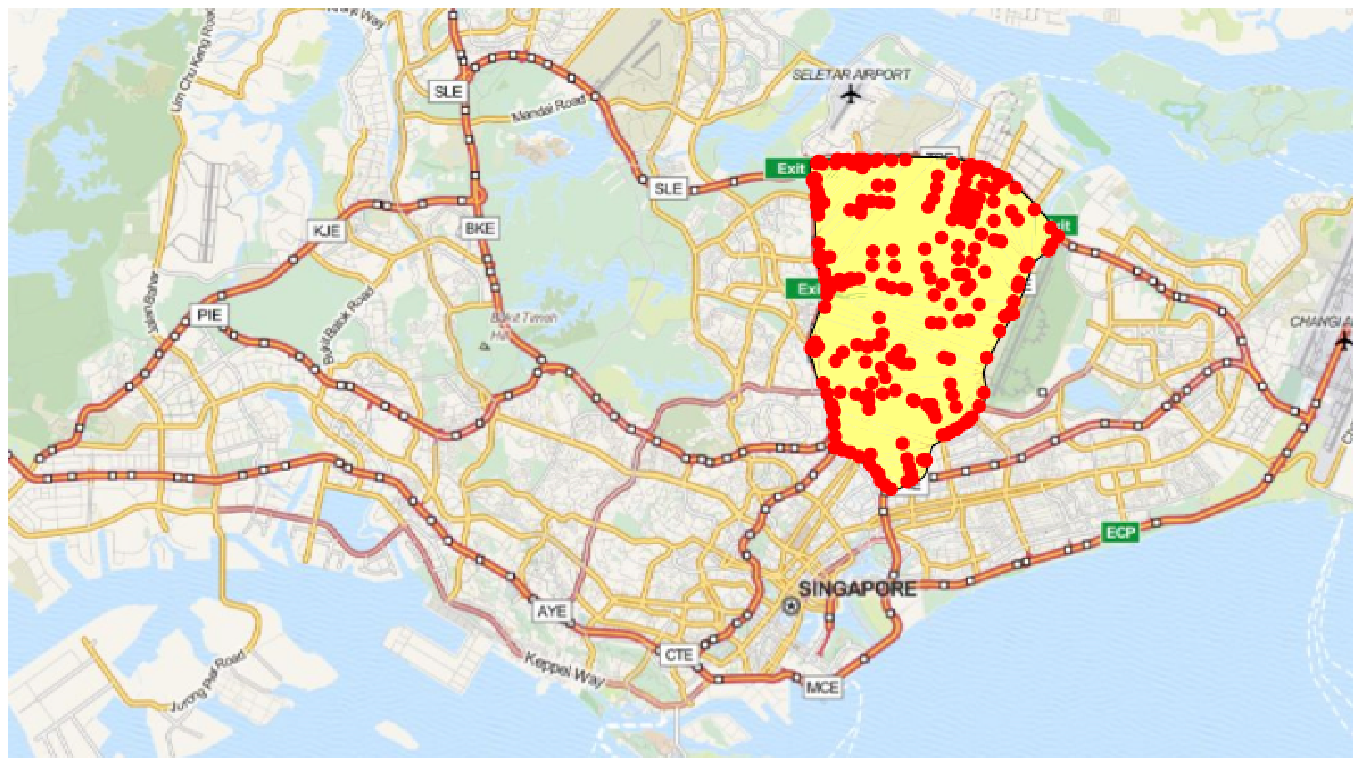}}
\subfloat[]{\includegraphics[width=0.3\textwidth]{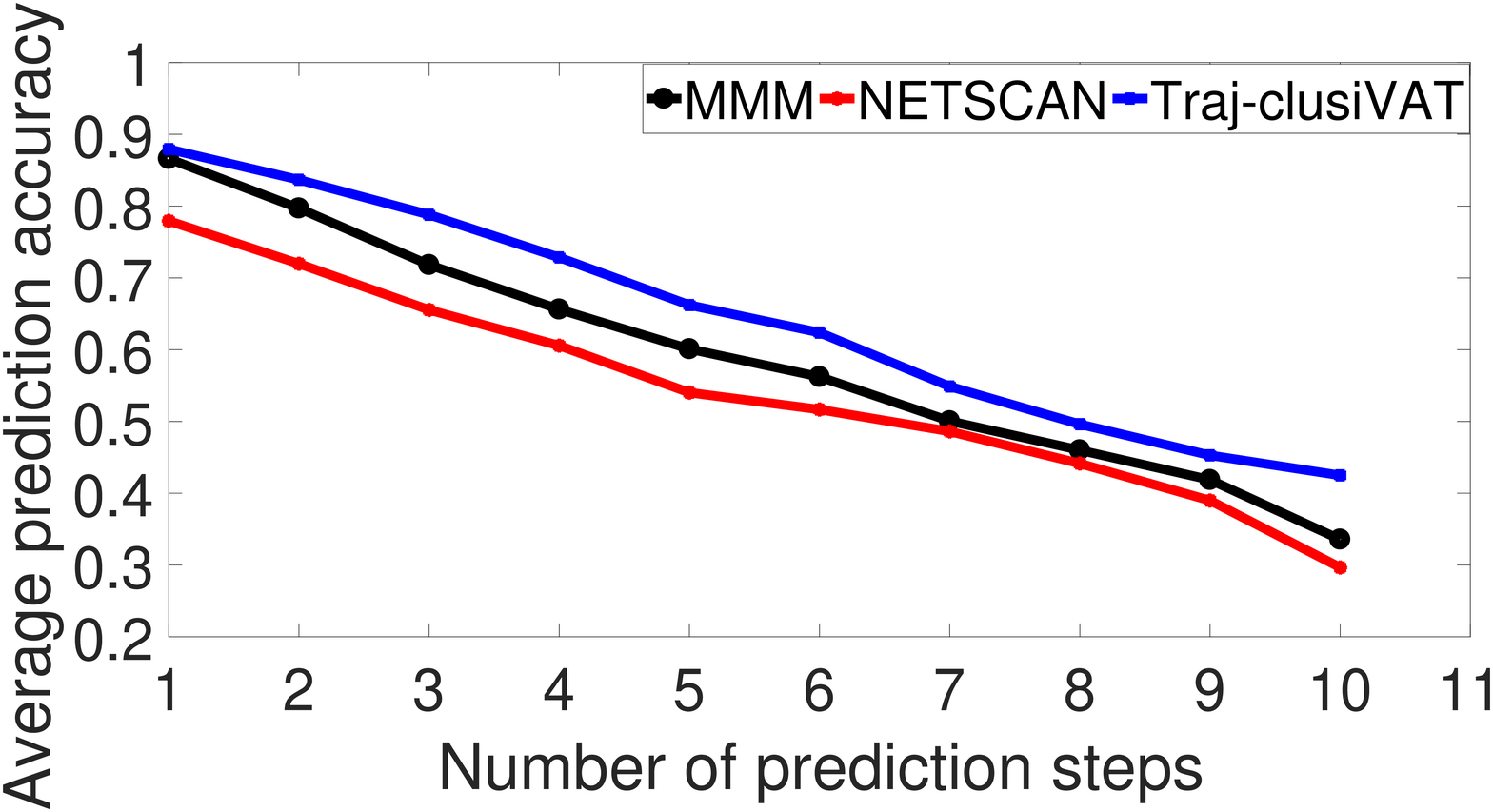}}
\subfloat[]{\includegraphics[width=0.3\textwidth]{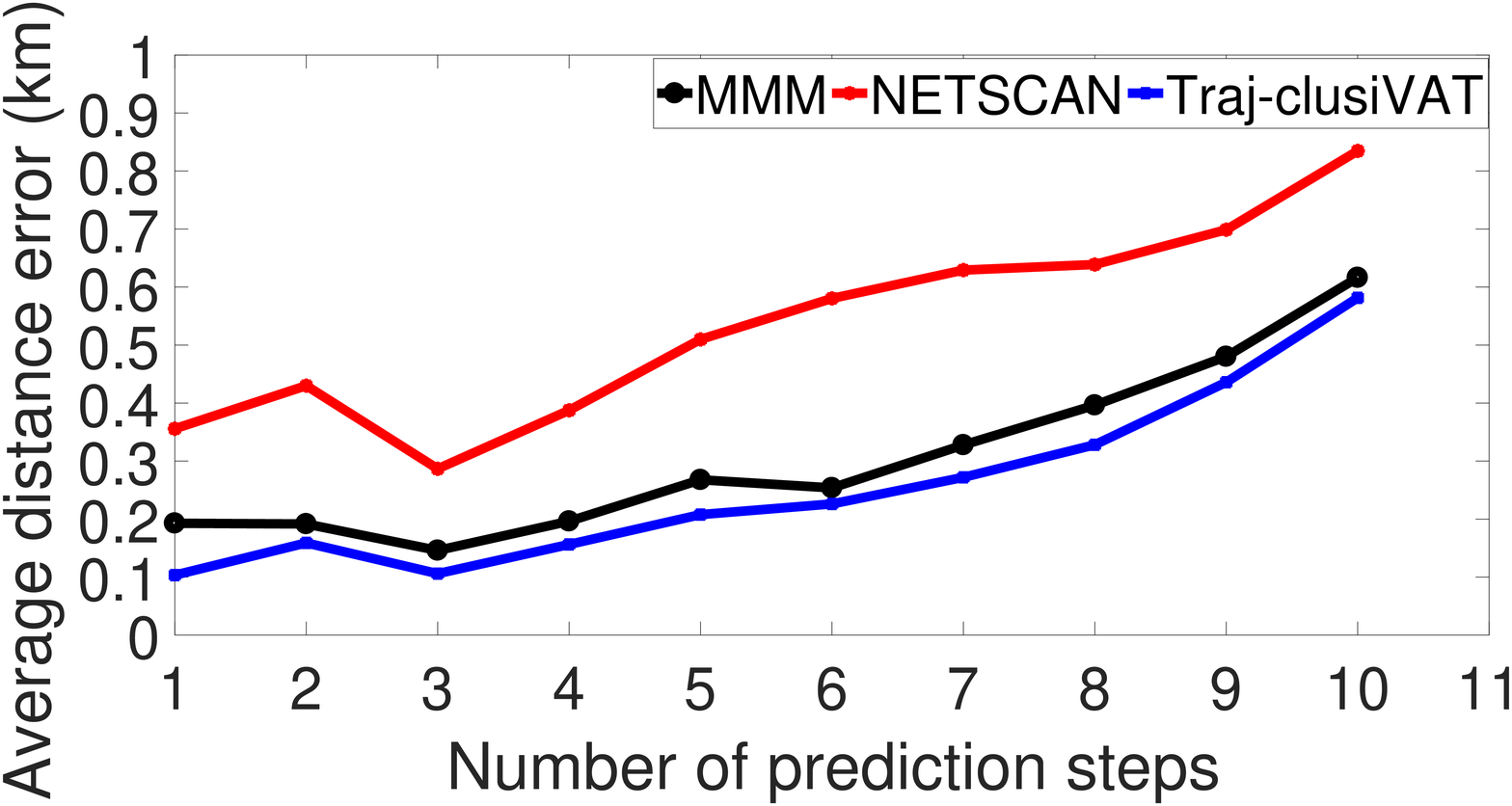}}
\caption{Singapore Taxi subset: (a) A small part (highlighted) of the Singapore road network; (b) Average prediction accuracy and (c) Average distance error comparison by prediction steps}
\label{Fig:DEvsPredTrajLengthSingaporeSubgraph}
\end{figure*}

In our experiments, we find that most of the clusters contain frequent trajectory patterns whose lengths are less than six or seven. Only a few clusters contain frequent trajectory patterns whose lengths are longer than seven steps. This finding conforms with the real-world situation, where a driver usually predicts only next few locations.

\begin{table}[]
\centering
\caption{Long-term prediction: Comparison of MMM, NETSCAN and Traj-clusiVAT}
\label{Table:LongtermPredictions}
\begin{tabular}{|l|c|c|c|}
\hline
\multicolumn{1}{|c|}{\multirow{2}{*}{}} & \multicolumn{3}{c|}{T-Drive Data} \\ \cline{2-4}
\multicolumn{1}{|c|}{} & Average PA & Average DE (km) & PR (\%) \\ \hline
MMM & $0.55$ & $0.68$ & $39.9$ \\ \hline
NETSCAN & $0.41$ & $0.87$ & $24.3$ \\ \hline
Traj-clusiVAT & $\textbf{0.62}$ & $\textbf{0.58}$ & $\textbf{49.8}$ \\ \hline
\hline
 & \multicolumn{3}{c|}{Singapore taxi data} \\ \hline
NETSCAN & $0.34$ & $1.41$ & $5.1$ \\ \hline
Traj-clusiVAT & $\textbf{0.59}$ & $\textbf{0.60}$ & $\textbf{24.8}$ \\ \hline
 & \multicolumn{3}{c|}{Singapore taxi subset} \\ \hline
MMM & $ 0.59$ & $ 0.30$ & $32$ \\ \hline
NETSCAN & $0.54$ & $0.53$ & $19$ \\ \hline
Traj-clusiVAT & $\textbf{0.64}$ & $\textbf{0.25}$ & $\textbf{46}$ \\ \hline
\end{tabular}
\end{table}

The average long-term prediction performance of all three approaches is summarized in Table~\ref{Table:LongtermPredictions}. The best performance is shown in bold for both datasets. Traj-clusiVAT achieves the highest PA, $0.62$ and $0.59$ and the lowest DE, $0.58$km and $0.60$km, for the T-Drive and Singapore taxi datasets, respectively. The MMM-based prediction approach is the second best method for T-Drive in terms of all three evaluation metrics. Traj-clusiVAT achieves prediction rates of $49.8\%$ and $24.8\%$ for the T-Drive and Singapore trajectory datasets, respectively. In other words, Traj-clusiVAT is able to predict complete trips for around $50\%$ of the trajectories in T-Drive, and for around $25\%$ of the trajectories in Singapore data. In contrast, MMM  predicts about $40\%$ of the total trajectories correctly for the T-Drive dataset. Although, NETSCAN performance improved for longer predictions due to longer dense paths, it only predicted about $5\%$ of the  total trajectories correctly. Overall, Traj-clusiVAT based prediction approach outperforms both MMM and NETSCAN-based prediction approaches based on all three evaluation metrics.

\textcolor{black}{As we could not apply MMM to the Singapore data due to its high computation and space complexity, we applied it to a subset of a small part of the Singapore road network, as shown in~\Fig~\ref{Fig:DEvsPredTrajLengthSingaporeSubgraph} (a). This sub-graph consists of $238$ nodes, $417$ edges, and  $828,870$ trajectories.  Similar to our previous experiments, we considered $60\%$ of the trajectories ($497,320$) randomly as training data and the remaining 40\% ($331,550$) as the test set, for both weekdays and weekend data. Although we could reduce the computational time and space requirements by considering a smaller road network, it still has high complexity due to large $N$ ($497,320$).  We could not run MMM with more than $10,000$ trajectories as it started to slow down our PC significantly and gave "Out of Memory" error message. Therefore, we used $10,000$ randomly selected trajectories as a training set for MMM-based TP to avoid associated computational and storage overload. We used the same subset for NETSCAN-TP and Traj-clusiVAT-TP for training in this experiment.}

\textcolor{black}{\Figs~\ref{Fig:DEvsPredTrajLengthSingaporeSubgraph} (b) (c) show the average prediction accuracy and average distance error, respectively, for all three algorithms. Traj-clusiVAT-TP method outperforms the MMM-based and NETSCAN-based TP approaches for this subset of the Singapore dataset, based on the average PA and DE. Overall, Traj-clusiVAT-TP achieves the highest average PA ($0.64$), lowest average DE ($0.25$km) and highest PR ($46$\%) among all three methods for Singapore taxi subset, as shown in Table~\ref{Table:LongtermPredictions}}. 

\begin{table}
\centering
\caption{Next location prediction:  Comparison of MMM, NETSCAN and Traj-clusiVAT}
\label{Table:NextLocationPrediction}
\scalebox{0.95}{
\begin{tabular}{|c|c|c|l|c|}
\hline
\multirow{2}{*}{}  & \multicolumn{2}{c|}{T-Drive} & \multicolumn{2}{c|}{Singapore Taxi} \\ \cline{2-5}
 & OA & ODE (km) & OA  & ODE (km)\\ \hline
MMM  & $0.77$  & $0.24$ &  - & - \\ \hline
NETSCAN & $0.67$  & $0.54$  & $0.62$ & $0.29$  \\ \hline
Traj-clusiVAT & $\textbf{0.80}$  & $\textbf{0.23}$ & $\textbf{0.86}$  & $\textbf{0.05}$  \\ \hline
\end{tabular}}
\end{table}

\subsection{Next location predictions}
In this experiment, we compare Traj-clusiVAT to the other two comparison approaches for predicting next locations. Given a taxi's current location, the next location prediction is to forecast the next location where the taxi may go. Table~\ref{Table:NextLocationPrediction} shows the one-step accuracy (OA) and one-step distance error (ODE) on the T-Drive and Singapore trajectory datasets. The Traj-clusiVAT-based approach predicts next location with more than $80\%$ accuracy and with distance error of less than a quarter of km for both T-Drive and Singapore data. The long-term prediction performance (Table~\ref{Table:LongtermPredictions}) of NETSCAN and Traj-clusiVAT is better for T-Drive than for the Singapore data. Conversely, the next location prediction performance of both approaches is better for the Singapore data than the T-Drive data. This may be because Singapore data contains a large number of longer trajectories that span entire Singapore city, whereas T-drive contains partial trajectories belonging to small part of the entire road network, hence modeling is not that efficient for T-drive data.
 In summary, Traj-clusiVAT outperforms both MMM and NETSCAN for next location prediction.

\subsection{Effect of latest locations of partial trajectory for prediction}
In the prediction step of Traj-clusiVAT, a partial trajectory $T^{p}=\{R_{1},R_{2},...,R_{l}\}$ is assigned to one of the $K$ clusters using our hybrid NPR approach. For a $T^{p}$, the best cluster is chosen based on either its path probability $P^{i}(T^{p})$ in each cluster or its trajDTW distance from each cluster (if $T^{p}$ is not fully contained in any cluster). The length of known partial trajectory $T^{p}$ increases after each next location prediction as $T^{p}$ is updated with a predicted location after each prediction, and subsequently, the updated $T^{p}$ is used for next location prediction, and so on.

We conduct an experiment in which instead of using full known partial trajectory $T^{p}$, we use only the latest movement steps or latest subsequence of $T^{p}$ until prediction to choose the best matching cluster in the hybrid NPR step. In this regard, we choose a different number of latest locations of known partial trajectories until prediction and investigate the effect on the performance for trajectory prediction.

\begin{figure}
\centering
\includegraphics[width=0.4\textwidth]{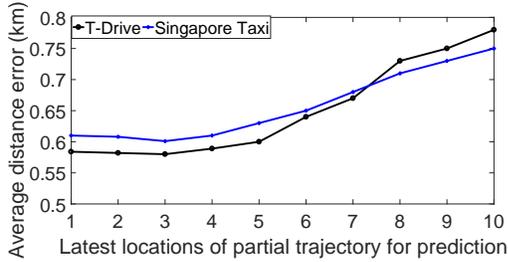}
\caption{Average DE vs latest locations of partial trajectory used to select best cluster in the hybrid NPR step.}
\label{Fig:AvePAvsLatestLocations}
\end{figure}

Fig~\ref{Fig:AvePAvsLatestLocations} shows the average distance error for a different number of latest locations of known partial trajectories until prediction for the T-drive and Singapore data. It can be inferred from the figure that the best performance is achieved when only the latest two or three locations of partial trajectory are used to find the best matching cluster. The average distance error increases if more than three latest locations are used to find the best cluster in the hybrid NPR step. This is because as the length of $T^{p}$ increases, its path probability in each cluster decreases, which means that the chance of sequence $T^{p}$ being fully contained in any cluster decreases.
  Moreover, if $T^{p}$ is not fully contained in any cluster representative trajectory, its distance from all the clusters increases with increasing length. This may result in wrong cluster assignment, which in turn, may degrade Traj-clusiVAT's prediction performance.

\subsection{Effect of Cut threshold $\alpha$}
\begin{figure}
\centering
\subfloat[T-Drive]{\includegraphics[width=0.248\textwidth]{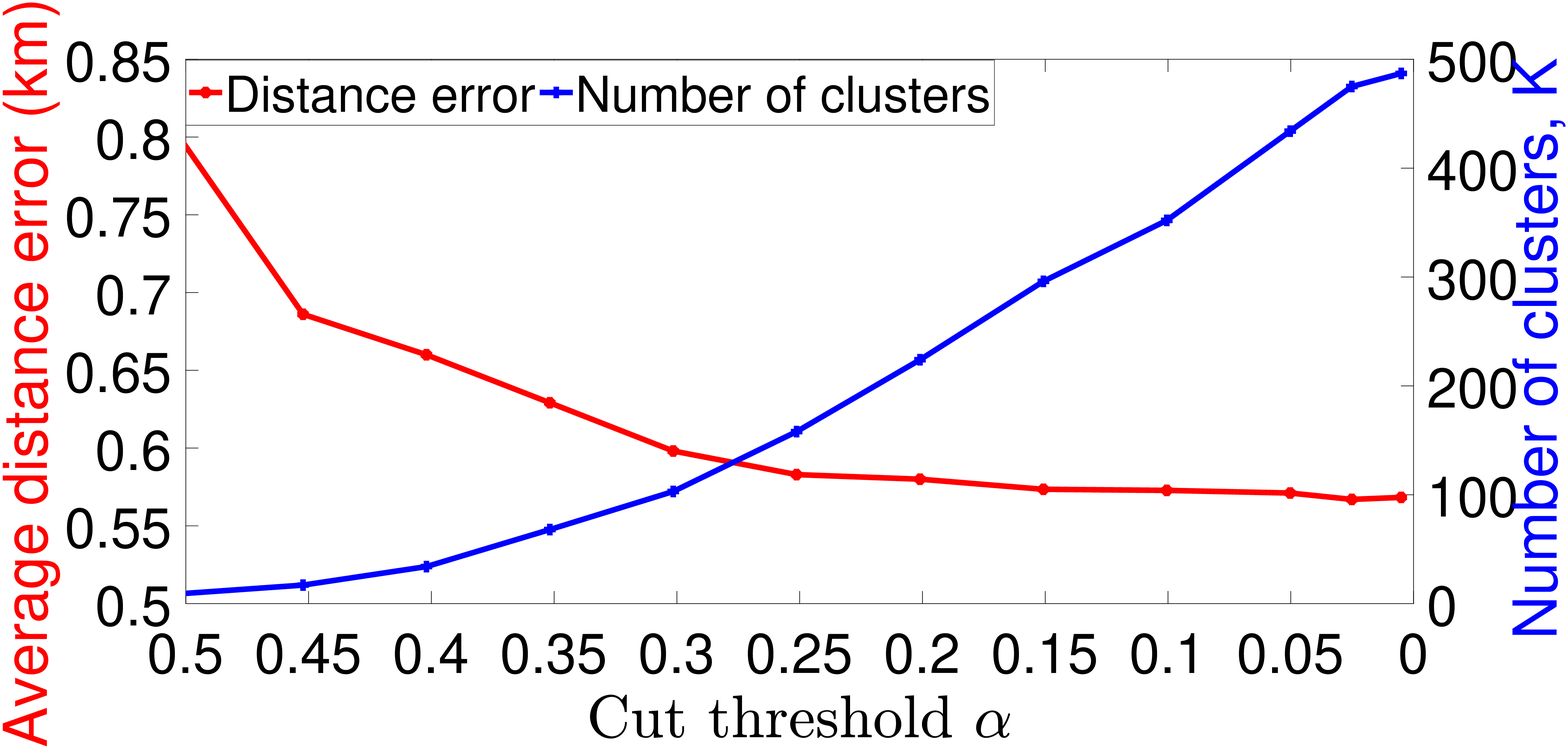}}
\subfloat[Singapore trajecory data]{\includegraphics[width=0.248\textwidth]{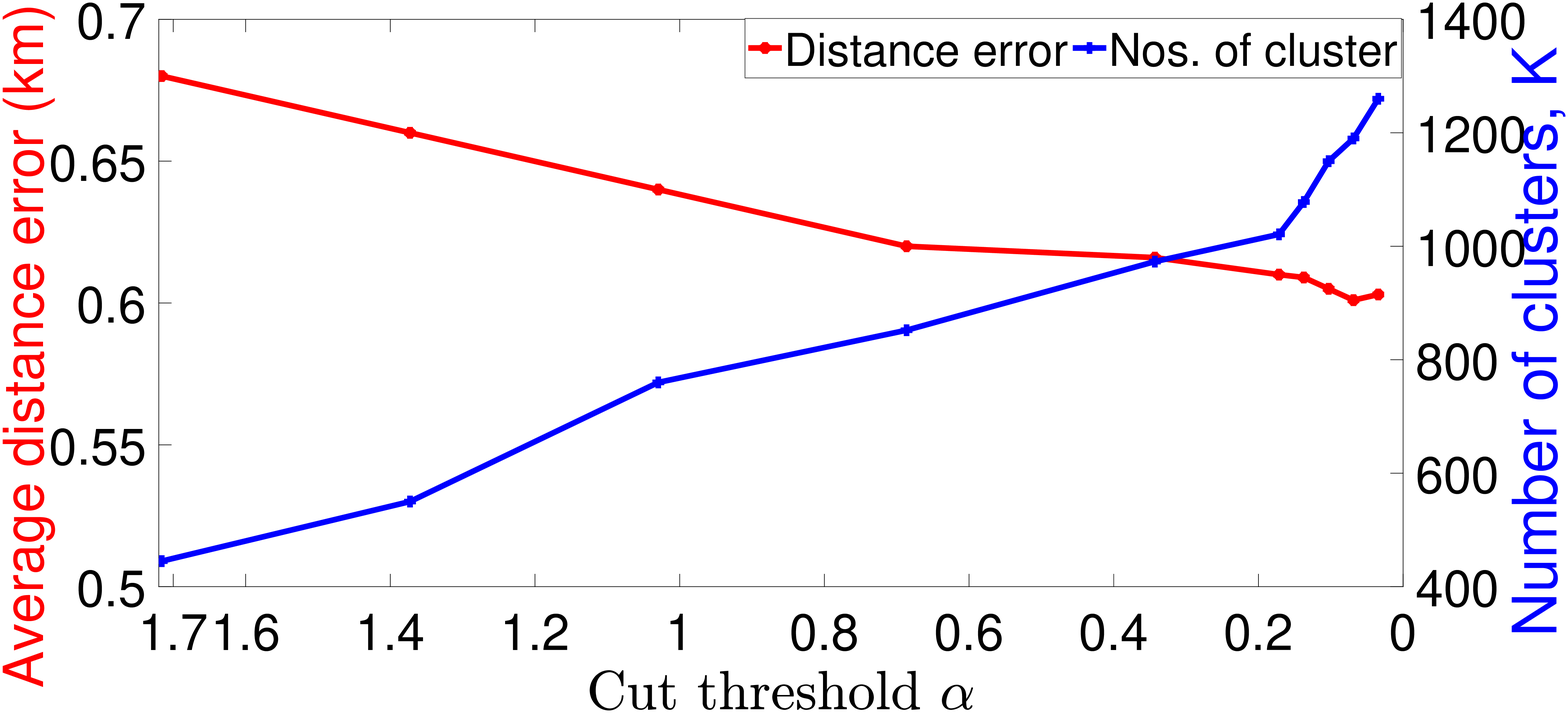}}
\caption{Effect of cut threshold $\alpha$}
\label{Fig:EffectofCutThreshold}
\end{figure}

In this experiment, we study the effect of cut threshold $\alpha$. The parameter $\alpha$ in Traj-clusiVAT controls how far two groups of data points should be from each other to be considered as different clusters. Figure~\ref{Fig:EffectofCutThreshold} shows the average DE and the number of clusters $K$ for different values of $\alpha$ for the T-Drive and Singapore data. The lower the cut threshold, the tighter the cluster boundaries, and hence, the higher the number of clusters. As the number of clusters $K$ increases, the average DE reduces. This is primarily because the higher $K$ corresponds to a larger number of unique frequent patterns, which improves the prediction performance. Figure~\ref{Fig:EffectofCutThreshold} shows that the Traj-clusiVAT performance improves with lower cut threshold $\alpha$ or with the higher number of clusters. However, with a large $K$, more MM needs to be trained, and hence, system complexity increases. 
 Moreover, Traj-clusiVAT performance does not improve significantly below a certain value of $\alpha$ for either dataset. The procedure to find an optimal value of $\alpha$ is described in~\cite{kumar2017visual}.

\subsection{Time performance analysis}
\begin{figure}[h]
\centering
\subfloat[T-Drive]{\includegraphics[width=0.26\textwidth,height=0.11\textheight]{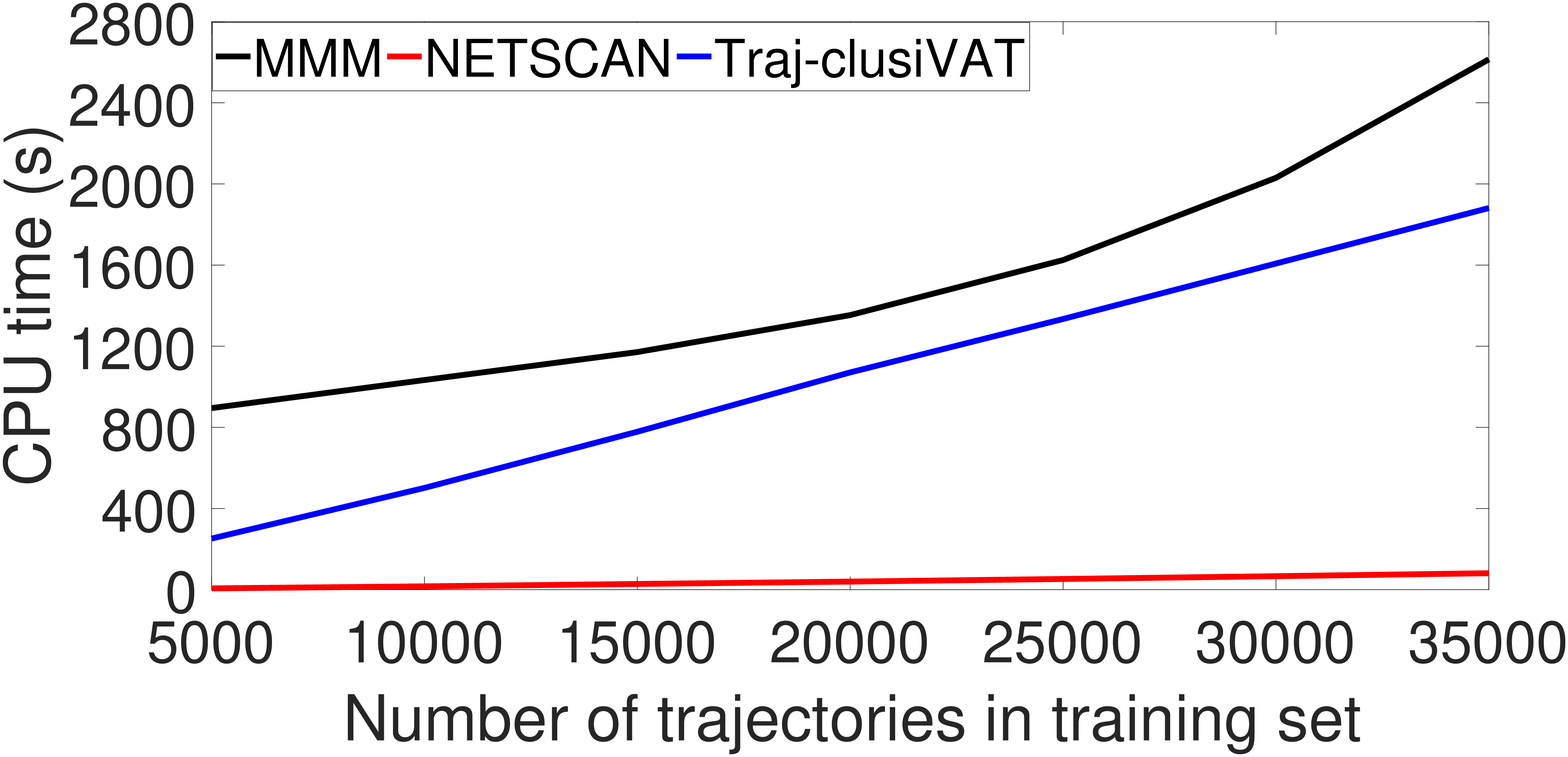}}
\subfloat[Singapore]{\includegraphics[width=0.26\textwidth,height=0.11\textheight]{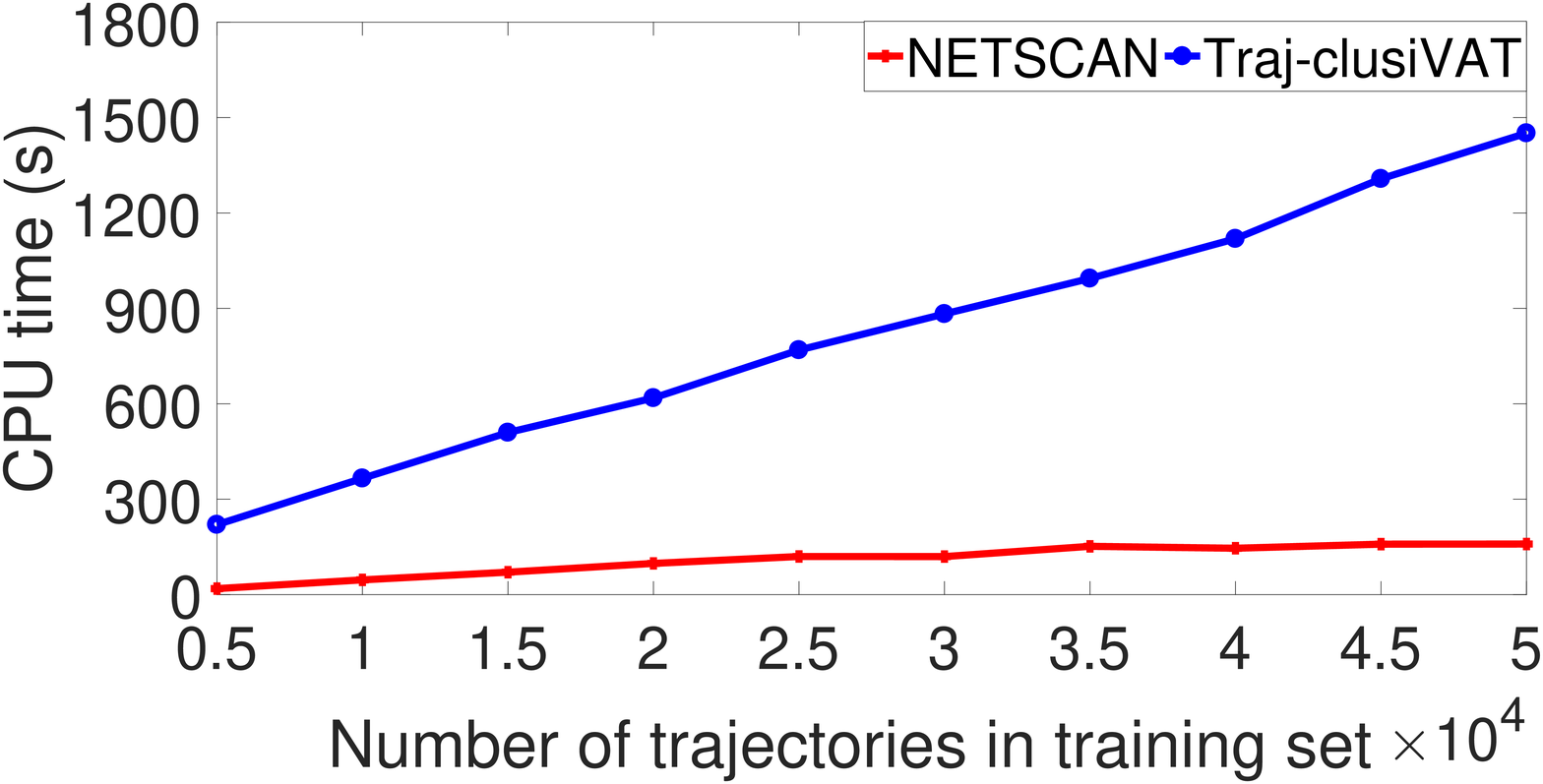}}
\caption{Training time comparison}
\label{Fig:TrainingTimeComparison}
\end{figure}

The training time of all three algorithms on different-size training sets is shown in~\Fig~\ref{Fig:TrainingTimeComparison}. The CPU-time for MMM increases most with the training data size because the computation of an intermediate matrix of size $|E| \times |E| \times N$ incurs high computational overhead and space complexity for large $N$. On the other hand, NETSCAN incurs the lowest computation time among all three methods. This is because it just computes dense paths based on the movement counts and density threshold, and assigns all trajectories to these dense paths based on similarity. Although it takes less time for training, it suffers from lower prediction accuracy. Traj-clusiVAT scales almost linearly in the number of trajectories, which make it scalable for big trajectory datasets.

Prediction-time is also an important criterion in real-time trajectory prediction. The average prediction time for all three approaches is presented in Table~\ref{Table:PredTimesTable}. We can see that all three approaches take similar times to forecast each trajectory for the T-drive dataset. The response time is less than $1.5$ms for T-drive, which suggests that all three approaches satisfy the requirement for real-time prediction. The average prediction time is higher for the Singapore dataset due to a large number of clusters identified by both NETSCAN and Traj-clusiVAT algorithms, but at $\sim 0.06$ seconds, it is negligible in terms of real-time prediction utility.

\begin{table}[h]
\centering
\caption{Prediction time in seconds for all three algorithms}
\label{Table:PredTimesTable}
\begin{tabular}{|l|l|l|l|}
\hline
 & MMM & NETSCAN & Traj-clusiVAT \\ \hline
T-Drive Taxi & $0.0014$s & $0.0011$s & $0.0012$s \\ \hline
Singapore Taxi & - & $0.063$s & $0.066$s \\ \hline
\end{tabular}
\end{table}

\section{Conclusions}\label{sec:conclusions}
Most existing TP approaches are not suitable for large volumes of overlapping trajectories in a dense road network. This article presents a novel, scalable,  hybrid architecture for short-term and long-term trajectory prediction, which can handle a large number of trajectories from a large-scale dense road network. The proposed framework is based on a scalable clustering approach, called Traj-clusiVAT, which is a modified version of clusiVAT for trajectory prediction. In particular, Traj-clusiVAT develops a novel algorithm to compute a representative trajectory for each cluster.  We also presented a new, hybrid nearest prototyping approach for accurate trajectory assignment to (one of) the clusters identified in previous steps of Traj-clusiVAT. Finally, we also propose a hybrid prediction framework based on hybrid NPR which can assign a query trajectory to best-matching cluster in a robust way to improve prediction performance.

We demonstrated the superiority of our proposed approach by comparing it with mixed Markov model and NETSCAN based TP approaches on two real trajectory datasets, including a large-scale trajectory dataset containing $3.28$ million trajectories of passenger trips obtained from $15,061$ taxis within Singapore over a period of one month. Our experimental results on both trajectory datasets show that Traj-clusiVAT based TP approach outperforms the other two approaches based on the prediction accuracy and distance error for short-term and long-term prediction for these two datasets. Our experimental results also suggest that Traj-clusiVAT satisfies the requirement for real-time predictions.  Our next effort will focus on online training in Traj-clusiVAT using incremental/decremental VAT approaches~\cite{kumar2016adaptive} to update clusters in real-time. We also intend to include additional factors such as speed, time, and user information in our prediction system to improve its prediction performance.
\section*{Acknowledgement}
We would like to thank 
  Institute for Infocomm Research (I2R), A*STAR Singapore for providing us with the Singapore taxi GPS log dataset.
\sloppy
\bibliographystyle{IEEEtran}
\bibliography{BibliographyTITS}

\begin{thebibliography}{10}
\providecommand{\url}[1]{#1}
\csname url@samestyle\endcsname
\providecommand{\newblock}{\relax}
\providecommand{\bibinfo}[2]{#2}
\providecommand{\BIBentrySTDinterwordspacing}{\spaceskip=0pt\relax}
\providecommand{\BIBentryALTinterwordstretchfactor}{4}
\providecommand{\BIBentryALTinterwordspacing}{\spaceskip=\fontdimen2\font plus
\BIBentryALTinterwordstretchfactor\fontdimen3\font minus
  \fontdimen4\font\relax}
\providecommand{\BIBforeignlanguage}[2]{{%
\expandafter\ifx\csname l@#1\endcsname\relax
\typeout{** WARNING: IEEEtran.bst: No hyphenation pattern has been}%
\typeout{** loaded for the language `#1'. Using the pattern for}%
\typeout{** the default language instead.}%
\else
\language=\csname l@#1\endcsname
\fi
#2}}
\providecommand{\BIBdecl}{\relax}
\BIBdecl

\bibitem{rathore2018real}
P.~Rathore, A.~S. Rao, S.~Rajasegarar, E.~Vanz, J.~Gubbi, and M.~Palaniswami,
  ``Real-time urban microclimate analysis using internet of things,''
  \emph{IEEE Internet of Things Journal}, vol.~5, no.~2, pp. 500--511, 2018.

\bibitem{zheng2008understanding}
Y.~Zheng, Q.~Li, Y.~Chen, X.~Xie, and W.-Y. Ma, ``Understanding mobility based
  on gps data,'' in \emph{Proceedings of the 10th international conference on
  Ubiquitous computing}.\hskip 1em plus 0.5em minus 0.4em\relax ACM, 2008, pp.
  312--321.

\bibitem{song2010limits}
C.~Song, Z.~Qu, N.~Blumm, and A.-L. Barab{\'a}si, ``Limits of predictability in
  human mobility,'' \emph{Science}, vol. 327, no. 5968, pp. 1018--1021, 2010.

\bibitem{lv2017big}
Q.~Lv, Y.~Qiao, N.~Ansari, J.~Liu, and J.~Yang, ``Big data driven hidden markov
  model based individual mobility prediction at points of interest,''
  \emph{IEEE Trans. on Vehicular Tech.}, vol.~66, no.~6, pp. 5204--5216, 2017.

\bibitem{kumar2016hybrid}
D.~Kumar, J.~C. Bezdek, M.~Palaniswami, S.~Rajasegarar, C.~Leckie, and T.~C.
  Havens, ``A hybrid approach to clustering in big data,'' \emph{IEEE
  Transactions on Cybernetics}, vol.~46, no.~10, pp. 2372--2385, 2016.

\bibitem{kumar2018fast}
D.~Kumar, H.~Wu, S.~Rajasegarar, C.~Leckie, S.~Krishnaswamy, and
  M.~Palaniswami, ``Fast and scalable big data trajectory clustering for
  understanding urban mobility,'' \emph{IEEE Transactions on Intelligent
  Transportation Systems}, no.~99, pp. 1--14, 2018.

\bibitem{bezdek2017book}
J.~C. Bezdek, \emph{{Primer on Cluster Analysis: Four Basic Methods that
  (Usually) Work}}.\hskip 1em plus 0.5em minus 0.4em\relax First Edition Design
  Publishing, 2017, vol.~1.

\bibitem{asahara2011pedestrian}
A.~Asahara, K.~Maruyama, A.~Sato, and K.~Seto, ``Pedestrian-movement prediction
  based on mixed markov-chain model,'' in \emph{Proceedings of the 19th ACM
  SIGSPATIAL international conference on advances in geographic information
  systems}.\hskip 1em plus 0.5em minus 0.4em\relax ACM, 2011, pp. 25--33.

\bibitem{kharrat2008clustering}
A.~Kharrat, I.~S. Popa, K.~Zeitouni, and S.~Faiz, ``Clustering algorithm for
  network constraint trajectories,'' in \emph{Headway in Spatial Data
  Handling}.\hskip 1em plus 0.5em minus 0.4em\relax Springer, 2008, pp.
  631--647.

\bibitem{morzy2007mining}
M.~Morzy, ``Mining frequent trajectories of moving objects for location
  prediction,'' in \emph{International Workshop on Machine Learning and Data
  Mining in Pattern Recognition}.\hskip 1em plus 0.5em minus 0.4em\relax
  Springer, 2007, pp. 667--680.

\bibitem{jeung2008hybrid}
H.~Jeung, Q.~Liu, H.~T. Shen, and X.~Zhou, ``A hybrid prediction model for
  moving objects,'' in \emph{IEEE 24th International Conference on Data
  Engineering (ICDE)}.\hskip 1em plus 0.5em minus 0.4em\relax Ieee, 2008, pp.
  70--79.

\bibitem{monreale2009wherenext}
A.~Monreale, F.~Pinelli, R.~Trasarti, and F.~Giannotti, ``Wherenext: a location
  predictor on trajectory pattern mining,'' in \emph{Proceedings of the 15th
  ACM SIGKDD international conference on Knowledge discovery and data
  mining}.\hskip 1em plus 0.5em minus 0.4em\relax ACM, 2009, pp. 637--646.

\bibitem{qiao2017predicting}
S.~Qiao, N.~Han, J.~Wang, R.-H. Li, L.~A. Gutierrez, and X.~Wu, ``Predicting
  long-term trajectories of connected vehicles via the prefix-projection
  technique,'' \emph{IEEE Trans. on Intell. Transport. Systems}, 2017.

\bibitem{ishikawa2004extracting}
Y.~Ishikawa, Y.~Tsukamoto, and H.~Kitagawa, ``Extracting mobility statistics
  from indexed spatio-temporal datasets.'' in \emph{STDBM}, 2004, pp. 9--16.

\bibitem{simmons2006learning}
R.~Simmons, B.~Browning, Y.~Zhang, and V.~Sadekar, ``Learning to predict driver
  route and destination intent,'' in \emph{Intelligent Transportation Systems
  Conference, 2006. ITSC'06. IEEE}.\hskip 1em plus 0.5em minus 0.4em\relax
  IEEE, 2006, pp. 127--132.

\bibitem{gambs2012next}
S.~Gambs, M.-O. Killijian, and M.~N. del Prado~Cortez, ``Next place prediction
  using mobility markov chains,'' in \emph{Proceedings of the First Workshop on
  Measurement, Privacy, and Mobility}.\hskip 1em plus 0.5em minus 0.4em\relax
  ACM, 2012, p.~3.

\bibitem{zhang2016gmove}
C.~Zhang, K.~Zhang, Q.~Yuan, L.~Zhang, T.~Hanratty, and J.~Han, ``Gmove:
  Group-level mobility modeling using geo-tagged social media,'' in
  \emph{Procd. of the 22nd ACM SIGKDD Intl. Conf. on Knowledge Discovery and
  Data Mining}, 2016, pp. 1305--1314.

\bibitem{altche2017lstm}
F.~Altch{\'e} and A.~De~La~Fortelle, ``An lstm network for highway trajectory
  prediction,'' in \emph{Intelligent Transportation Systems (ITSC), 2017 IEEE
  20th International Conference on}.\hskip 1em plus 0.5em minus 0.4em\relax
  IEEE, 2017, pp. 353--359.

\bibitem{wu2017modeling}
H.~Wu, Z.~Chen, W.~Sun, B.~Zheng, and W.~Wang, ``Modeling trajectories with
  recurrent neural networks.''\hskip 1em plus 0.5em minus 0.4em\relax IJCAI,
  2017.

\bibitem{bock2017self}
J.~Bock, T.~Beemelmanns, M.~Kl{\"o}sges, and J.~Kotte, ``Self-learning
  trajectory prediction with recurrent neural networks at intelligent
  intersections,'' 2017.

\bibitem{yuan2017review}
G.~Yuan, P.~Sun, J.~Zhao, D.~Li, and C.~Wang, ``A review of moving object
  trajectory clustering algorithms,'' \emph{Artificial Intelligence Review},
  vol.~47, no.~1, pp. 123--144, 2017.

\bibitem{won2009trajectory}
J.-I. Won, S.-W. Kim, J.-H. Baek, and J.~Lee, ``Trajectory clustering in road
  network environment,'' in \emph{IEEE Symposium on Comp. Intelligence and Data
  Mining, 2009.}\hskip 1em plus 0.5em minus 0.4em\relax IEEE, 2009, pp.
  299--305.

\bibitem{lee2007trajectory}
J.-G. Lee, J.~Han, and K.-Y. Whang, ``Trajectory clustering: a
  partition-and-group framework,'' in \emph{Proceedings of the 2007 ACM SIGMOD
  Intl. Conf. on Management of data}.\hskip 1em plus 0.5em minus 0.4em\relax
  ACM, 2007, pp. 593--604.

\bibitem{wang2012clustering}
Y.~Wang, Q.~Han, and H.~Pan, ``A clustering scheme for trajectories in road
  networks,'' in \emph{Advanced Technology in Teaching-Proceedings of the 2009
  3rd International Conference on Teaching and Computational Science (WTCS
  2009)}.\hskip 1em plus 0.5em minus 0.4em\relax Springer, 2012, pp. 11--18.

\bibitem{roh2010nncluster}
G.-P. Roh and S.-w. Hwang, ``Nncluster: An efficient clustering algorithm for
  road network trajectories,'' in \emph{Intl. Conf. on Database Systems for
  Advanced Applications}.\hskip 1em plus 0.5em minus 0.4em\relax Springer,
  2010, pp. 47--61.

\bibitem{han2015road}
B.~Han, L.~Liu, and E.~Omiecinski, ``Road-network aware trajectory clustering:
  Integrating locality, flow, and density,'' \emph{IEEE Transactions on Mobile
  Computing}, vol.~14, no.~2, pp. 416--429, 2015.

\bibitem{ashbrook2003using}
D.~Ashbrook and T.~Starner, ``Using gps to learn significant locations and
  predict movement across multiple users,'' \emph{Personal and Ubiquitous
  computing}, vol.~7, no.~5, pp. 275--286, 2003.

\bibitem{mathew2012predicting}
W.~Mathew, R.~Raposo, and B.~Martins, ``Predicting future locations with hidden
  markov models,'' in \emph{Proceedings of the 2012 ACM conference on
  ubiquitous computing}.\hskip 1em plus 0.5em minus 0.4em\relax ACM, 2012, pp.
  911--918.

\bibitem{chen2015predicting}
M.~Chen, Y.~Liu, and X.~Yu, ``Predicting next locations with object clustering
  and trajectory clustering,'' in \emph{Pacific-Asia Conference on Knowledge
  Discovery and Data Mining}.\hskip 1em plus 0.5em minus 0.4em\relax Springer,
  2015, pp. 344--356.

\bibitem{ying2011semantic}
J.~J.-C. Ying, W.-C. Lee, T.-C. Weng, and V.~S. Tseng, ``Semantic trajectory
  mining for location prediction,'' in \emph{Proceedings of the 19th ACM
  SIGSPATIAL International Conference on Advances in Geographic Information
  Systems}.\hskip 1em plus 0.5em minus 0.4em\relax ACM, 2011, pp. 34--43.

\bibitem{wiest2012probabilistic}
J.~Wiest, M.~H{\"o}ffken, U.~Kre{\ss}el, and K.~Dietmayer, ``Probabilistic
  trajectory prediction with gaussian mixture models,'' in \emph{Intelligent
  Vehicles Symposium (IV), 2012 IEEE}.\hskip 1em plus 0.5em minus 0.4em\relax
  IEEE, 2012, pp. 141--146.

\bibitem{lei2011exploring}
P.-R. Lei, T.-J. Shen, W.-C. Peng, and J.~Su, ``Exploring spatial-temporal
  trajectory model for location prediction,'' in \emph{12th IEEE Intl. Conf. on
  Mobile Data Management (MDM)}, vol.~1.\hskip 1em plus 0.5em minus 0.4em\relax
  IEEE, 2011, pp. 58--67.

\bibitem{chen2010system}
L.~Chen, M.~Lv, and G.~Chen, ``A system for destination and future route
  prediction based on trajectory mining,'' \emph{Pervasive and Mobile
  Computing}, vol.~6, no.~6, pp. 657--676, 2010.

\bibitem{Kumar2015ASF}
D.~Kumar, S.~Rajasegarar, M.~Palaniswami, X.~Wang, and C.~Leckie, ``A scalable
  framework for clustering vehicle trajectories in a dense road network,'' in
  \emph{The 4th International Workshop on Urban Computing (UrbComp), Held in
  conjunction with the 21th ACM SIGKDD}, 2015.

\bibitem{bezdek2002vat}
J.~C. Bezdek and R.~J. Hathaway, ``{VAT}: A tool for visual assessment of
  (cluster) tendency,'' in \emph{Proceedings of the International Joint
  Conference on Neural Networks}, vol.~3.\hskip 1em plus 0.5em minus
  0.4em\relax IEEE, 2002, pp. 2225--2230.

\bibitem{havens2012efficient}
T.~C. Havens and J.~C. Bezdek, ``An efficient formulation of the improved
  visual assessment of cluster tendency (ivat) algorithm,'' \emph{IEEE Trans.
  on Knlwdg. and Data Engg.}, vol.~24, no.~5, pp. 813--822, 2012.

\bibitem{hathaway2006scalable}
R.~J. Hathaway, J.~C. Bezdek, and J.~M. Huband, ``Scalable visual assessment of
  cluster tendency for large data sets,'' \emph{Pattern Recognition}, vol.~39,
  no.~7, pp. 1315--1324, 2006.

\bibitem{rathore2018approximate}
P.~Rathore, J.~C. Bezdek, D.~Kumar, S.~Rajasegarar, and M.~Palaniswami,
  ``Approximate cluster heat maps of large high-dimensional data,'' in
  \emph{24th International Conference on Pattern Recognition (ICPR)}.\hskip 1em
  plus 0.5em minus 0.4em\relax IEEE, 2018, pp. 195--200.

\bibitem{kumar2017visual}
D.~Kumar, J.~C. Bezdek, S.~Rajasegarar, C.~Leckie, and M.~Palaniswami, ``A
  visual-numeric approach to clustering and anomaly detection for trajectory
  data,'' \emph{The Visual Computer}, vol.~33, no.~3, pp. 265--281, 2017.

\bibitem{yavacs2005data}
G.~Yava{\c{s}}, D.~Katsaros, {\"O}.~Ulusoy, and Y.~Manolopoulos, ``A data
  mining approach for location prediction in mobile environments,'' \emph{Data
  \& Knowledge Engineering}, vol.~54, no.~2, pp. 121--146, 2005.

\bibitem{lee2008traclass}
J.-G. Lee, J.~Han, X.~Li, and H.~Gonzalez, ``Traclass: trajectory
  classification using hierarchical region-based and trajectory-based
  clustering,'' \emph{Procd. of the VLDB Endowment}, vol.~1, no.~1, pp.
  1081--1094, 2008.

\bibitem{lu2011mining}
E.~H.-C. Lu, V.~S. Tseng, and S.~Y. Philip, ``Mining cluster-based temporal
  mobile sequential patterns in location-based service environments,''
  \emph{IEEE transactions on knowledge and data engineering}, vol.~23, no.~6,
  pp. 914--927, 2011.

\bibitem{sung2012trajectory}
C.~Sung, D.~Feldman, and D.~Rus, ``Trajectory clustering for motion
  prediction,'' in \emph{Intelligent Robots and Systems (IROS), 2012 IEEE/RSJ
  International Conference on}.\hskip 1em plus 0.5em minus 0.4em\relax IEEE,
  2012, pp. 1547--1552.

\bibitem{ferreira2013vector}
N.~Ferreira, J.~T. Klosowski, C.~E. Scheidegger, and C.~T. Silva, ``Vector
  field k-means: Clustering trajectories by fitting multiple vector fields,''
  in \emph{Computer Graphics Forum}, vol.~32, no. 3pt2.\hskip 1em plus 0.5em
  minus 0.4em\relax Wiley Online Library, 2013, pp. 201--210.

\bibitem{barbehenn1998note}
M.~Barbehenn, ``A note on the complexity of dijkstra's algorithm for graphs
  with weighted vertices,'' \emph{IEEE transactions on computers}, vol.~47,
  no.~2, p. 263, 1998.

\bibitem{salvador2007toward}
S.~Salvador and P.~Chan, ``Toward accurate dynamic time warping in linear time
  and space,'' \emph{Intelligent Data Analysis}, vol.~11, no.~5, pp. 561--580,
  2007.

\bibitem{yuan2010t}
J.~Yuan, Y.~Zheng, C.~Zhang, W.~Xie, X.~Xie, G.~Sun, and Y.~Huang, ``T-drive:
  driving directions based on taxi trajectories,'' in \emph{Proceedings of the
  18th SIGSPATIAL International conference on advances in geographic
  information systems}.\hskip 1em plus 0.5em minus 0.4em\relax ACM, 2010, pp.
  99--108.

\bibitem{lu2015taxi}
Y.~Lu, S.~Xiang, and W.~Wu, ``Taxi queue, passenger queue or no queue?'' in
  \emph{Proc. of 18th International Conference on Extending Database Technology
  (EDBT). Brussels, Belgium}, 2015, pp. 593--604.

\bibitem{GraphHopper}
``{GraphHopper, "Map-matching"},'' {http://www.unhabitat.org/pmss/listItem}
  {Details.aspx?publicationID=3387}, 2017.

\bibitem{newson2009hidden}
P.~Newson and J.~Krumm, ``Hidden markov map matching through noise and
  sparseness,'' in \emph{Proceedings of the 17th ACM SIGSPATIAL international
  conference on advances in geographic information systems}.\hskip 1em plus
  0.5em minus 0.4em\relax ACM, 2009, pp. 336--343.

\bibitem{besse2017destination}
P.~C. Besse, B.~Guillouet, J.-M. Loubes, and F.~Royer, ``Destination prediction
  by trajectory distribution-based model,'' \emph{IEEE Transactions on
  Intelligent Transportation Systems}, 2017.

\bibitem{huang2017mining}
Q.~Huang, ``Mining online footprints to predict user's next location,''
  \emph{International Journal of Geographical Information Science}, vol.~31,
  no.~3, pp. 523--541, 2017.

\bibitem{qiao2015traplan}
S.~Qiao, N.~Han, W.~Zhu, and L.~A. Gutierrez, ``Traplan: an effective
  three-in-one trajectory-prediction model in transportation networks,''
  \emph{IEEE Transactions on Intelligent Transportation Systems}, vol.~16,
  no.~3, pp. 1188--1198, 2015.

\bibitem{jeung2007mining}
H.~Jeung, H.~T. Shen, and X.~Zhou, ``Mining trajectory patterns using hidden
  markov models,'' in \emph{International Conference on Data Warehousing and
  Knowledge Discovery}.\hskip 1em plus 0.5em minus 0.4em\relax Springer, 2007,
  pp. 470--480.

\bibitem{kumar2016adaptive}
D.~Kumar, J.~C. Bezdek, S.~Rajasegarar, M.~Palaniswami, C.~Leckie, J.~Chan, and
  J.~Gubbi, ``Adaptive cluster tendency visualization and anomaly detection for
  streaming data,'' \emph{ACM Transactions on Knowledge Discovery from Data
  (TKDD)}, vol.~11, no.~2, p.~24, 2016.

\end{thebibliography}

\end{document}